\documentclass[accepted]{uai2026} 

\usepackage{microtype}
\usepackage{graphicx}
\usepackage{float}
\usepackage{subcaption}
\usepackage{booktabs}
\usepackage{multirow}
\usepackage{adjustbox}
\usepackage{placeins}

\usepackage{amsmath,amssymb,amsthm,mathtools}
\usepackage{bbm}

\usepackage{enumitem}
\usepackage{xcolor}

\usepackage{algorithm}
\usepackage{algpseudocode}

\usepackage[numbers,sort&compress]{natbib}

\newtheorem{theorem}{Theorem}[section]
\newtheorem{lemma}[theorem]{Lemma}
\newtheorem{corollary}[theorem]{Corollary}
\newtheorem{proposition}[theorem]{Proposition}

\theoremstyle{definition}
\newtheorem{definition}[theorem]{Definition}
\newtheorem{remark}[theorem]{Remark}
\newtheorem{example}[theorem]{Example}

\newtheorem{condition}[theorem]{Condition}

\DeclareMathOperator{\E}{\mathbb{E}}
\DeclareMathOperator{\Var}{Var}

\newcommand{\PPUO}{\mathrm{P\textsc{puo}}}

\usepackage[american]{babel}
\usepackage{tikz}

\title{Fast Best-in-Class Regret for Contextual Bandits}

\author[1]{Samuel~Girard}
\author[2]{Aur\'elien~Bibaut}
\author[1]{Jill-Jênn Vie}
\author[3]{Arthur~Gretton}
\author[2, 4]{Nathan~Kallus}
\author[3]{Houssam~Zenati}
\affil[1]{INRIA Soda}
\affil[2]{Netflix}
\affil[3]{Gatsby Computational Neuroscience Unit, University College London}
\affil[4]{Cornell University and Cornell Tech}

\begin{document}
\maketitle

\begin{abstract}
We study the problem of stochastic contextual bandits in the agnostic setting, where the goal is to compete with the best policy in a given class without assuming realizability or imposing model restrictions on losses or rewards. In this work, we propose \textit{Online Pessimistic Policy Learning} and establish the first fast rate for regret relative to the best-in-class policy.
Our proposed algorithm updates the policy at every round by minimizing a pessimistic objective, defined as a clipped inverse-propensity estimate of the policy value plus a variance penalty. By leveraging entropy assumptions on the policy class and a H\"olderian error-bound condition (a generalization of the margin condition), we achieve fast best-in-class regret rates, including polylogarithmic rates in the parametric case. Our analysis is driven by a novel sequential self-normalized maximal inequality for bounded martingale empirical processes, which yields uniform variance-adaptive confidence bounds and guarantees pessimism under adaptive data collection.
\end{abstract}

\section{Introduction}

Personalizing decisions to individuals is crucial across many user-facing domains, from improving health outcomes in medicine to increasing revenue in online advertising. In a new decision-making environment, one must learn how to personalize while making decisions that minimize loss, or maximize reward. This is the contextual bandit setting: contexts, representing individual information, arrive sequentially, and at each time the learner must choose an action, balancing good present decisions with learning for the future. In the stochastic variant, these instances are drawn iid from a fixed but unknown joint distribution over contexts and potential losses, one for each action.

A primary aim is sublinear regret, so that long-run time average loss converges to the average attainable if no learning were needed. To achieve sublinear regret, the literature on stochastic contextual bandits generally addresses two settings: value-based or policy-based. In the value-based setting, we assume the context-conditional expected loss function belongs to a given function class, whether linear \citep{abbasi2011improved,li2010contextual,rusmevichientong2010linearly,goldenshluger2013linear,bastani2020online}, non-parametric \citep{rigollet2010nonparametric,perchet2013multi,hu2020smooth}, or a generic class with a regression oracle \citep{foster2020beyond,foster2020instance,foster2021efficient,simchi2022bypassing,pacchiano2024second}. Regret in this setting corresponds to comparing against the personalization rule that minimizes the context-conditional expected loss.

Notably, regret rates in the value-based setting can be significantly sped up by accounting for the \textit{margin condition} satisfied by the environment distribution \citep{goldenshluger2013linear,bastani2020online,bastani2021mostly,rigollet2010nonparametric,perchet2013multi,hu2020smooth}:
the less mass the context distribution places on hard contexts—those where the best and second-best actions have nearly equal loss—the faster the attainable regret.


In the policy-based setting, we are given a class of policies $\Pi$, make no realizability assumptions on the  loss model, and instead consider competing with the best policy in the class \citep{dudik2011efficient,agarwal2014tamingmonsterfastsimple,foster2018contextual,bibaut2020generalized}. This approach is \textit{agnostic} and \textit{model-free}, imposing no assumptions on the context-loss joint distribution and competing with what is achievable inside the policy class.
However, unlike value-based approaches, existing policy-based bandit algorithms do not enjoy \emph{fast} best-in-class rates under low-noise / margin-like assumptions.

In this paper, we introduce \textit{Online Pessimistic Policy Learning} (O2PL) and provide the first \textit{fast} rates for best-in-class regret in stochastic contextual bandits in the model-agnostic policy-based setting.
Our speedup depends on a H\"olderian error bound (HEB), which generalizes the margin condition beyond the realizable setting.
Our algorithm performs pessimistic policy learning at every round, minimizing
over $\Pi$ a clipped inverse-propensity estimate of policy value plus a
variance-aware penalty. We do not impose \emph{uniform} overlap---a floor
$\pi(a\mid x)\ge\eta>0$ that would bound the importance ratios outright;
clipping instead controls large ratios adaptively.
Our main technical contribution is a novel uniform, Bernstein-type maximal
inequality for martingale empirical processes with bounded increments,
localized by the \emph{realized} rather than the predictable quadratic
variation, which yields confidence radii that adapt to an empirical (clipped)
importance-weight variance proxy. We expect this inequality to be of
independent interest. The one quantity we leave \emph{unclipped} is the
importance-weight variance in the HEB, since it is precisely what governs the clipping bias.

\paragraph{Organization.} Section~\ref{sec:related} reviews related work, and Section~\ref{sec:setup} formalizes the agnostic best-in-class contextual bandit setting and assumptions. Section~\ref{sec:algorithm} presents O2PL, our clipped IPW algorithm with a variance-adaptive pessimistic penalty, and states its regret guarantee. Section~\ref{sec:master} gives a master theorem for any algorithm satisfying abstract surrogate-risk, bias, and variance conditions; Section~\ref{sec:analysis} then verifies these conditions for O2PL, using our new self-normalized maximal inequality. Section~\ref{sec:numerics} reports experiments.

\section{Related literature}
\label{sec:related}

We focus on the policy-based, best-in-class strand of contextual bandits.
In adversarial settings, expert aggregation and exponential weights \citep{vovk1990aggregating,littlestone1994weighted,auer2002nonstochastic} led to Exp4-style algorithms and chaining-based nonparametric online learning \citep{dudley1967sizes, cesa2017algorithmic,chatterji2019online, martin2022nested}.
In stochastic settings, oracle-efficient methods progressed from epoch-greedy \citep{langford2008epoch} to regret-optimal approaches based on cost-sensitive classification oracles \citep{dudik2011efficient,agarwal2014tamingmonsterfastsimple,foster2018practical,foster2020beyond,foster2018contextual}.

Off-policy evaluation relies on inverse propensity weighting (IPW) \citep{thompson1952,rosenbaum1983} and refinements such as doubly robust estimators \citep{robins1995semiparametric,dudik2011,dudik2014,wang2017optimal}.
Policy learning optimizes such estimators over a policy class, often with normalization or variance control \citep{swaminathan2012,joachims2018deep,KallusNIPS2018,kallus2021optimal,kallus2022doubly,AtheyWager2021,ZhouAtheyWager2022,zenati23scrm,jin2025policylearningwithoutoverlap,haddouche2025logsmooth}.
For adaptively collected logs, risk guarantees were established by \citet{Kallus2021,zhan2021policy}.

Our analysis is rooted in maximal inequalities from empirical-process theory \citep{dudley1967sizes,van1996weak,vandeGeer2000}.
For sequential dependence, Freedman-type martingale Bernstein inequalities \citep{freedman1975tail} are central.
We require a \emph{uniform} self-normalized inequality over policy classes under adaptive sampling. Our development is closest in spirit to sequential complexity tools \citep{rakhlin2015sequential} and modern anytime self-normalized inference \citep{waudbysmith2022estimatingmeansboundedrandom,anytime_ian}, but adapted to clipped importance-weight losses and logit-scale entropy.

\section{Problem Setup}
\label{sec:setup}
We consider the stochastic contextual bandit setting:
at each round $t=1,\dots,T$, the learner observes a context $X_t \in \mathcal X$, selects a policy $\pi_t(\cdot\mid X_t) \in \Pi$, samples an action $A_t\sim \pi_t(\cdot\mid X_t)$, and incurs a loss $Y_t\in[-1,0]$. We denote $Z_t=(X_t,A_t,Y_t)$, and let $\mathcal{F}_{t-1} := \sigma(Z_1,\ldots,Z_{t-1})$ be the sigma-field induced by $Z_1,\ldots,Z_{t-1}$. 
We take the context space $\mathcal X$ to be a standard Borel space, the action set $\mathcal A$ to be finite or standard Borel, and policies $\pi(\cdot\mid x)$ to be measurable stochastic kernels on $\mathcal A$. Contexts are i.i.d.\ draws $X_t\sim P_X$ from a fixed distribution $P_X$ on $\mathcal X$, and losses are generated by a fixed conditional kernel, so that the law of $Y_t\mid X_t=x,A_t=a$ is independent of $t$; we write $\mu(x,a):=\mathbb E[Y\mid X=x,A=a]$ for its conditional mean. The policy $\pi_t$ is $\mathcal{F}_{t-1}$-measurable, i.e.\ determined only by data observed so far.

We are given a policy class $\Pi$ and consider the cumulative pseudo-regret against the best-in-class policy:
\begin{equation}
\mathrm{Regret}_T \:=\ \sum_{t=1}^T \Big( R(\pi_t) - \inf_{\pi\in\Pi}R(\pi) \Big),
\label{eq:regret_def}
\end{equation}
where
\begin{equation}
    R(\pi) = \mathbb{E}_{X\sim P_X} \mathbb{E}_{A \sim \pi(\cdot \mid X)} \mu(X,A),
\end{equation}
and we write $\pi^\star \in \arg\min_{\pi\in\Pi} R(\pi)$ for a best-in-class policy.

To obtain sublinear regret, we impose structure on the policy class $\Pi$ together with a low-noise condition relating excess risk to the variance of importance weights; we do not require any realizability assumption on $\mu$. The range $[-1,0]$ is only a normalization: any bounded loss interval is admissible, with constants scaling in the range; we do not treat heavy-tailed or unbounded losses.

\section{Online Pessimistic Policy Learning}
\label{sec:algorithm}
We now present our algorithm, \textit{Online Pessimistic Policy Learning} (O2PL). The design principle is simple: at each round, optimize a lower confidence bound on a centered policy value, where uncertainty is measured by an empirical variance proxy. This yields conservative updates when coverage is poor, and aggressive updates when the data support a policy well.

Define the weight-truncated IPW counterfactual loss:
\begin{align} \label{eq:clipped_loss}
    \ell_{t,\alpha}(\pi)(z)
    := \left\{ \min\left( \frac{\pi(a \mid x)}{\pi_t(a \mid x)}, \alpha \right) - 1 \right\} y.
\end{align}
We will later choose the truncation parameter $\alpha$ to grow with $t$ to ensure truncation introduces only a vanishing bias.

Given this, define the centered clipped surrogate target
\begin{align}
    R_{t,\alpha}(\pi)
    &:= \frac{1}{t} \sum_{s=1}^t \mathbb E\!\left[\ell_{s,\alpha}(\pi)(Z_s)\mid \mathcal F_{s-1}\right], \\
    R_t(\pi)
    &:= \frac{1}{t} \sum_{s=1}^t \mathbb E\!\left[\left(\frac{\pi}{\pi_s}(A_s \mid X_s)-1\right)Y_s \mid \mathcal F_{s-1}\right].
\end{align}
The unclipped centered surrogate differs from the true risk only by an additive baseline:
\[
R_t(\pi) = R(\pi) - \frac{1}{t}\sum_{s=1}^t R(\pi_s),
\]
so that $R_t(\pi)-R_t(\pi^\star)=R(\pi)-R(\pi^\star)$.

We estimate this centered surrogate with the empirical clipped objective
\begin{align}
    \widehat R_{t,\alpha}(\pi) &:= \frac{1}{t} \sum_{s=1}^t \ell_{s, \alpha}(\pi)(Z_s), \label{eq:empirical_risk}
\end{align}
and penalize using the clipped empirical variance proxy
\begin{align}
    \widehat \sigma_{t,\alpha}^2(\pi) &:= \frac{1}{t} \sum_{s=1}^t \left( \min\!\left(\alpha, \frac{\pi}{\pi_s} (A_s \mid X_s)\right) - 1\right)^2 \label{eq:empirical_std}
\end{align}

For the theoretical analysis, we also define the unclipped predictable variance
\begin{align}
    \sigma_s^2(\pi)
    &:= \mathbb E\!\left[\left( \frac{\pi}{\pi_s}(A_s\mid X_s)-1\right)^2 \,\middle|\, \mathcal F_{s-1}\right],
    \\
    \bar \sigma_t^2(\pi)
    &:= \frac1t\sum_{s=1}^t \sigma_s^2(\pi),
\end{align}
and abbreviate $\sigma_s^2 := \sigma_s^2(\pi^\star)$, $\bar\sigma_t^2 := \bar\sigma_t^2(\pi^\star)$.
The unclipped variance $\bar\sigma_t^2$ is the natural quantity appearing in the H\"olderian error bound (Condition~\ref{asm:HEB}) and in the clipping bias analysis, while the clipped proxy $\widehat\sigma_{t,\alpha}$ is what the algorithm can compute and what controls the empirical process.

O2PL relies at each time $t$ on a pessimistic policy-update oracle $\PPUO_t$: it returns the policy
minimizing an upper confidence bound on the centered risk, namely the
empirical estimate $\widehat R_t$ plus a variance-aware penalty $\Phi_p$.
\begin{definition}[Optimization oracle] The function
    $\PPUO_t: (0,\infty) \times (0,1) \times \mathcal Z^{\times t} \times \Pi^{\times t} \to \Pi$ is a pessimistic policy update oracle that maps a clipping level $\alpha > 0$, a confidence level $\delta \in (0,1)$, and a sequence of observations $Z_1,\ldots,Z_t \in \mathcal Z$ and policies $\pi_1,\ldots,\pi_t \in \Pi$ to an element of
    \begin{equation}
        \arg\min_{\pi \in \Pi} \widehat R_t(\pi) + \Phi_p(\widehat \sigma_{t,\alpha}(\pi), \alpha, t, \delta),
    \end{equation}
    where
\[
\begin{aligned}
\Phi_p(\sigma,\alpha,t,\delta)
&:= \left(\frac{\alpha}{t}\right)^{2/(2+p)}
   + \frac{\sigma^{1-p/2}}{\sqrt t} \\
&\quad + \sigma\sqrt{\frac{\log(1/\delta)}{t}}
   + \frac{\alpha\log(1/\delta)}{t}.
\end{aligned}
\]
    with the convention that $\sigma^{1-p/2} \equiv \sigma \cdot \mathrm{polylog}(e/\sigma)$ when $p = 0$.
\end{definition}
\begin{algorithm}
\caption{Online Pessimistic Policy Learning (O2PL)}\label{alg:OPPL}
\begin{algorithmic}
\Require $T \geq 1$, $p \in [0,2)$, $\beta > 0$, $\pi_0 \in \Pi$, $\delta \in (0,1)$, $(\PPUO_t)_{t=1}^T$.
\State Initialize $\pi_1 \gets \pi_0$.
\For{$t = 1,\ldots,T$}
    \State Set $\alpha_t$ as in Theorem~\ref{thm:main_regret},
    \State $\delta_t \gets \delta / (t(t+1))$,
    \State Receive context $X_t$, sample $A_t \sim \pi_t( \cdot \mid X_t)$, receive loss $Y_t$,
    \State $\pi_{t+1} \gets \PPUO_t(\alpha_t, \delta_t, (Z_1,\ldots,Z_t), (\pi_1,\ldots,\pi_t))$,
\EndFor
\end{algorithmic}
\end{algorithm}
\subsection{Main regret bound}

We now present the main bound on the regret of O2PL to the best in class. Our result depends on the following two conditions which are discussed further in Section \ref{sec:discussion_entropy} and \ref{sec:discussion_HEB}.

\begin{condition}[Covering entropy of policy class on the logit scale]\label{asm:entropy_logPi_scale}
    There exists $p\in[0,2)$ such that for every $\epsilon \in (0,1]$,
    \(
        \log N(\epsilon, \log \Pi, \| \cdot \|_\infty)
        \lesssim
        \log(e/\epsilon)
    \)
    if $p=0$, and $\lesssim \epsilon^{-p}$ if $p\in(0,2)$.
\end{condition}

\begin{condition}[H\"olderian policy error bound]\label{asm:HEB}
There exists $\beta \in (0,1]$ such that for every $\pi \in \Pi$,
\begin{align}
    \Var_{X \sim P_X, A \sim \pi( \cdot \mid X)}\left( \frac{\pi^\star(A \mid X)}{\pi(A \mid X)} \right) \lesssim \left(R(\pi) - R(\pi^\star)\right)^\beta.
\end{align}
\end{condition}

\begin{theorem}
\label{thm:main_regret}
    Suppose that Conditions~\ref{asm:entropy_logPi_scale} and ~\ref{asm:HEB} hold. Then the sequence of policies generated by Algorithm~\ref{alg:OPPL} satisfies with probability at least $1-\delta$:
    if $\beta = 1$, for $\alpha_t := 1+\log(et)$,
    \begin{align}
    \mathrm{Regret}_T \;=\;
    \begin{cases}
      \mathcal{O}(\log^2 T) & \text{for } p = 0,\\[0.3em]
      \widetilde{\mathcal{O}}\!\left(T^{\frac{p}{2+p}}\right) & \text{for } p \in (0,2),
    \end{cases}
    \end{align}
    and, if $0 < \beta < 1$, for $\alpha_t := 2 \vee t^\gamma$ with
    $\gamma := \frac{2(1 -\beta )}{2(2 - \beta) + p }$,
    \begin{align}
    \mathrm{Regret}_T = \widetilde{\mathcal{O}} \left(
    T^{\frac{2(1 - \beta) + p }{2(2 - \beta) + p }}
    \right),
    \end{align}
where the $\widetilde{\mathcal{O}}$ notation absorbs polylogarithmic factors in $T$ and in $1 / \delta$ (including the parametric $p=0$ entropy logarithms).
\end{theorem}

The theorem does not require any overlap assumption and formalizes the central message of the paper: under entropy control and a H\"olderian error-bound condition, policy-based agnostic contextual bandits admit fast best-in-class rates, with polylogarithmic regret in the parametric case.

\paragraph{On the overlap assumption.}
When the learner requires uniform overlap, $\pi_t(a\mid x)\ge \eta>0$ for all $t,a,x$, the importance ratios are bounded by $\eta^{-1}$, so one may fix $\alpha_t\equiv\eta^{-1}$; clipping is then inactive and the surrogate coincides with the unbiased centered risk. Writing $\bar\Delta_t := \tfrac1t\sum_{s=1}^t (R(\pi_s)-R(\pi^\star))$, the H\"olderian bound and Jensen give $\bar\sigma_t^2\lesssim\bar\Delta_t^\beta$, and the resulting fixed-point relation solves to $\mathrm{Regret}_T = \widetilde{\mathcal O}(T^{\frac{2(1-\beta)+\beta p}{2(2-\beta)+\beta p}})$.

This is exactly the classical fast-rate exponent for full-information ERM under margin/error-bound conditions \citep{MassartNedelec2006RiskBounds}: identifying $r=p/2$, $\theta=1/\beta$ recovers $n^{-1/(2-\beta+\beta p/2)}$, reducing to $n^{-2/(2+p)}$ when $\beta=1$ and to Stone's rate $n^{-2k/(2k+d)}$ for smooth classes with $p=d/k$ \citep{Stone1982,RakhlinSridharanTsybakov2017EmpiricalEntropy}. Thus, under overlap, partial monitoring carries no additional polynomial price. Without overlap one must let $\alpha_t\to\infty$ to control clipping bias, and this extra bias--variance tradeoff is exactly what yields the slower exponent in Theorem~\ref{thm:main_regret}.

\subsection{Discussion of the policy class entropy condition}
\label{sec:discussion_entropy}
Condition~\ref{asm:entropy_logPi_scale} quantifies the size of the policy class through a single exponent $p$:
\[
\log N(\epsilon,\log\Pi,\|\cdot\|_\infty)\lesssim
\begin{cases}
\log(e/\epsilon), & p=0,\\
\epsilon^{-p}, & p\in(0,2).
\end{cases}
\]
Smaller $p$ means lower complexity and therefore faster rates in Theorem~\ref{thm:main_regret}. This is the same role played by metric-entropy exponents in classical empirical-process theory \citep{dudley1967sizes,van1996weak,vandeGeer2000}. Typical examples are:
\begin{itemize}
    \item Finite or finite-dimensional parametric policy classes (including compact softmax/logit models with uniformly Lipschitz logits in parameters): $\log N(\epsilon,\log\Pi,\|\cdot\|_\infty)=O(\log(1/\epsilon))$, so $p=0$.
    \item H\"older logit classes $\log\Pi\subset C^s([0,1]^d)$ with bounded $C^s$-norm: $\log N(\epsilon,\log\Pi,\|\cdot\|_\infty)\asymp \epsilon^{-d/s}$, so $p=d/s$.
    \item Sobolev/Mat\'ern-type smooth classes on $d$-dimensional contexts (with enough smoothness for uniform control): the same heuristic scaling $p\approx d/s$ is standard.
\end{itemize}

The log-scale is significant. Our losses and confidence bounds depend on policy \emph{ratios} $\pi/\pi_t=\exp(\log\pi-\log\pi_t)$, so uniform control of $\log\pi$ gives multiplicative control of importance ratios. This is exactly the right geometry for variance-sensitive policy learning.

It also requires the policy class to behave well near the boundary. If a class allows $\pi(a\mid x)=0$, it implies $\log\pi(a\mid x)=-\infty$, and $\log\Pi$ may have infinite sup-norm entropy near that boundary. In practice, one works with interior policies (e.g., softmax with temperature/flooring, $\epsilon$-greedy mixtures, or explicit propensity floors), which keeps $\log\Pi$ well-behaved and simultaneously stabilizes importance-weight variance.

\subsection{Discussion of the H\"olderian policy error bound}
\label{sec:discussion_HEB}

Condition~\ref{asm:HEB} introduced by \citet{zenati23scrm} is the structural assumption that drives fast rates in our analysis. It links a \emph{variance/overlap} quantity to suboptimality:
\[
\Var\left(\frac{\pi^\star(A\mid X)}{\pi(A\mid X)}\right)
=
\E_X\!\left[\chi^2\!\left(\pi^\star(\cdot\mid X)\,\|\,\pi(\cdot\mid X)\right)\right],
\]
so the condition says that as a policy gets close in risk to $\pi^\star$, its local mismatch with $\pi^\star$ (measured in $\chi^2$-divergence) must also shrink at a polynomial rate. This is exactly the quantity that controls the variance of importance-weighted evaluation, so the condition directly explains why pessimistic updates can accelerate.

This condition is a policy-learning analogue of H\"olderian error bounds in optimization, where one controls a geometry term (distance to the solution set, gradient norm, KL-type divergence) by a power of objective suboptimality. In particular, for smooth objectives, the Polyak--\L ojasiewicz (PL) inequality $\|\nabla f(\theta)\|^2 \gtrsim f(\theta)-f^\star$ is one canonical instance of this geometry and yields linear-type convergence. In that literature, such bounds underlie restart-based acceleration and fast rates \citep{polyak1963,lojasiewicz1963,lojasiewicz1993,bolte2007,becker2011,nesterov2012GradientMF,daspremont21}; see also related online-learning uses \citep{gaillard_2018}.

\citet{zenati23scrm} discuss instances of losses and policy classes for which this condition holds. At a high level, Condition~\ref{asm:HEB} should hold whenever two ingredients are present: (i) low excess risk forces near-agreement with the optimal policy on most contexts, and (ii) policy parameterization enforces enough overlap so that the $\chi^2$-mismatch is controlled by disagreement. The next lemma gives a concrete realizable-margin instantiation.

\begin{lemma}[Realizability and Tsybakov margin imply HEB]
\label{lem:margin_implies_heb}
Suppose the action space $\mathcal A=\{1,\dots,K\}$ is finite. Define the conditional mean $\mu(x,a):=\E[Y\mid X=x,A=a]$, the optimal value $\mu^\star(x):=\min_{a}\mu(x,a)$, and the optimal action selector $a^\star(x)\in\arg\min_a\mu(x,a)$. Assume the following hold:
\begin{enumerate}[label=(\roman*), leftmargin=*]
    \item \textbf{Realizability:} There exists $\pi^\star\in\Pi$ such that $R(\pi^\star)=\E[\mu^\star(X)]$.
    \item \textbf{Tsybakov margin condition \citep{mammen}:} Let the margin gap be $\Delta(x):=\min_{a\neq a^\star(x)}(\mu(x,a)-\mu^\star(x))$. There exist constants $C_{\mathrm m}>0$ and $\kappa>0$ such that $\Pr(\Delta(X)\le u)\le C_{\mathrm m}u^\kappa$ for all $u\ge0$.
    \item \textbf{Optimal-action overlap:} There exists a constant $c_0>0$ such that $\pi(a^\star(X)\mid X)\ge c_0$ almost surely for all $\pi\in\Pi$.
\end{enumerate}
Then, there exists a constant $C>0$ such that for all $\pi\in\Pi$,
\[
    \Var_{X,\,A\sim\pi(\cdot\mid X)}\!\left(\frac{\pi^\star(A\mid X)}{\pi(A\mid X)}\right)
    \le
    C\big(R(\pi)-R(\pi^\star)\big)^{\frac{\kappa}{\kappa+1}}.
\]
In particular, Condition~\ref{asm:HEB} holds with $\beta=\kappa/(\kappa+1)$.
\end{lemma}

Lemma~\ref{lem:margin_implies_heb}, whose proof is given in Appendix~\ref{app:realizability}, formalizes the main intuition: margins convert risk suboptimality into small disagreement mass on near-tie contexts, and overlap converts this disagreement mass into low importance-weight variance. This is precisely the mechanism exploited by our sequential pessimistic updates. 

The previous lemma is meant as a structural example for the variance--error relation itself; realizability is not necessary for Condition~\ref{asm:HEB}. Moreover, combining it simultaneously with the log-scale entropy condition typically requires working with a smoothed or floored version of the optimal policy. In practice, constructions such as $\epsilon$-greedy or softmax policies with temperature/flooring are common ways to enforce the required optimal-action overlap.

\begin{example}[No Realizability Instantiation]
\label{rem:heb_no_realizability}
Take a single context, two actions with mean losses $\mu_0,\mu_1$ and gap $\Delta:=\mu_0-\mu_1>0$, and the floored class $\Pi=\{\pi_\theta:\pi_\theta(1)=\theta,\ \theta\in[c,1-c]\}$, $c\in(0,\tfrac12)$. The Bayes policy is deterministic and excluded, so realizability fails; the best-in-class policy is $\pi^\star=\pi_{1-c}$. Then $R(\pi_\theta)-R(\pi^\star)=(1-c-\theta)\Delta$ while:
\[
\Var_{A\sim\pi_\theta}(\pi^\star(A)/\pi_\theta(A))=(1-c-\theta)^2/(\theta(1-\theta)),
\]
which is quadratic in the excess risk. Hence Condition~\ref{asm:HEB} holds with $\beta=1$, the fastest regime, despite non-realizability.
\end{example}

\section{A master theorem for agnostic fast rates}
\label{sec:master}

This section isolates the reusable proof template behind the regret analysis. Rather than committing to a specific estimator, we state abstract conditions that separate three roles: (i) control of surrogate excess risk by pessimism, (ii) control of surrogate-to-true-risk bias, and (iii) a variance-to-excess-risk link. The key abstraction is the \emph{critical radius}, which quantifies the resolution limit of the surrogate risk estimator over the policy class.

Consider a sequence of policies $(\pi_t)_{t \geq 1} \in \Pi^{\mathbb{N}}$. For every $t \geq 1$ and $\alpha > 0$, let $R_{t,\alpha} : \Pi \to \mathbb{R}$ be a surrogate risk functional and $\widehat R_{t,\alpha}$ its empirical counterpart. Let $(\sigma_t)_{t \geq 1}$, $(\alpha_t)_{t \geq 1}$ be positive sequences, and write $\bar\sigma_t := (t^{-1} \sum_{s=1}^t \sigma_s^2)^{1/2}$. It is helpful to think of $R_{t,\alpha}$ as a clipped surrogate risk, $\alpha_t$ as a bias--complexity trade-off parameter, and $\sigma_t^2$ as a predictable variance proxy.

\paragraph{Critical radius.}
The accuracy of $\widehat R_{t,\alpha}$ as a uniform estimate of $R_{t,\alpha}$ over $\Pi$ depends on the complexity of the surrogate loss class. We capture this through a localized sequential complexity measure $\mathcal H_t(\cdot, \alpha) : (0,\infty) \to [0,\infty)$, which quantifies the metric entropy of the loss class localized to functions with variance at most $r$, relative to sample size $t$ and envelope $\alpha$. The \emph{critical radius} is the smallest scale at which this complexity is dominated by a quadratic signal term:
\begin{equation}\label{eq:critical_radius_abstract}
  r_{t,\alpha}
  := \inf\!\left\{
    r > 0 \;:\; \mathcal H_t(r, \alpha) \le r^2
  \right\}.
\end{equation}
Below $r_{t,\alpha}$, sampling noise overwhelms differences between surrogate risks; above it, uniform inference over the policy class becomes possible.

\begin{condition}[Excess surrogate risk]\label{asm:surrogate_risk}
There exist $b \in [0,1]$ and a decreasing sequence $(\delta_t)_{t \geq 1}$ in $(0,1)$ such that for every $t \geq 1$, with probability at least $1-\delta_t$,
\begin{align}
R_{t, \alpha_t}(\pi_{t+1}) - R_{t, \alpha_t}(\pi^\star)
&\lesssim
\frac{\bar\sigma_t^{1-b}}{\sqrt{t}} + r_{t,\alpha}^2,
\label{eq:surrogate_excess}
\end{align}
where the critical radius satisfies $r_{t,\alpha} \asymp (\alpha_t / t)^{1/(2+2b)}$.
\end{condition}

The parameter $b$ plays the role of a complexity exponent: $b=0$ corresponds to parametric classes and $b = p/2 \in (0,1)$ to nonparametric classes with entropy exponent $p$. The condition says the pessimistic update controls surrogate excess risk up to a variance-adaptive term $\bar\sigma_t^{1-b}/\sqrt{t}$ that shrinks as the importance-weight variance decreases, and a critical-radius floor $r_{t,\alpha}^2$ set by class complexity and clipping level.

\begin{condition}[Surrogate risk bias]\label{asm:surrogate_bias}
For every $t \geq 1$,
\begin{align}
\left(R(\pi_{t+1}) - R(\pi^\star)\right)
&- \left(
    R_{t,\alpha_t}(\pi_{t+1})
    - R_{t,\alpha_t}(\pi^\star)
\right) \lesssim \frac{1}{\alpha_t} \bar\sigma_t^2.
\label{eq:surrogate_bias}
\end{align}
\end{condition}

\begin{condition}[Variance bound]\label{asm:heb}
There exists $\beta\in(0,1]$ such that, for every $t\ge 1$,
$\sigma_t^2 \lesssim (R(\pi_t)-R(\pi^\star))^\beta$.
\end{condition}

\begin{remark}[Bias--complexity tradeoff]
Conditions~\ref{asm:surrogate_risk} and~\ref{asm:surrogate_bias} encode a tension controlled by $(\alpha_t)$. The critical-radius floor $r_{t,\alpha_t}^2$ increases with $\alpha$, since a larger envelope enlarges the localized complexity $\mathcal H_t(r,\alpha_t)$ and pushes the critical radius up; the surrogate bias $\bar\sigma_t^2/\alpha_t$ shrinks with $\alpha_t$. Condition~\ref{asm:heb} breaks this deadlock: by linking $\bar\sigma_t^2$ to excess risk, it ensures the bias term decays along the trajectory, allowing $\alpha_t$ to grow slowly enough that the critical radius stays controlled. The proof of Theorem~\ref{thm:regret} chooses $\alpha_t$ to equalize the two competing rates.
Notably, the variance-localization term $\bar\sigma_t^{\,1-b}/\sqrt t$---the
engine of the fast rate in full-information ERM---never binds here: because
the running variance $\bar\sigma_t$ shrinks with the excess risk along the
trajectory, it decays at least as fast as the balanced rate (equal only when
$b=0$ or $\beta=1$). The rate is set entirely by the critical radius against the clipping bias, so the H\"olderian bound accelerates through the bias channel rather than through
localization, in contrast to the overlap regime, where clipping is inactive
and localization alone sets the rate.
\end{remark}

\begin{theorem}[Regret bound]\label{thm:regret}
Suppose Conditions~\ref{asm:surrogate_risk}--\ref{asm:heb} hold. Then, with probability at least $1-\delta$, if $\beta = 1$, for $\alpha_t := 1+\log(et)$, $\delta_t=\delta/(t(t+1))$,
    \begin{align}
    \mathrm{Regret}_T \;=\;
    \begin{cases}
      \mathcal{O}(\log^2 T) & \text{for } b = 0,\\[0.3em]
      \widetilde{\mathcal{O}}\!\left(T^{\frac{b}{1+b}}\right) & \text{for } b \in (0,1],
    \end{cases}
    \end{align}
    and, if $0 < \beta < 1$, for $\alpha_t := 2 \vee t^\gamma$ with $\gamma := \frac{1-\beta}{2+b-\beta}$,
    \begin{align}
    \mathrm{Regret}_T = \widetilde{\mathcal{O}} \left(
    T^{\frac{1 - \beta + b}{2 - \beta + b}}
    \right),
    \end{align}
where $\widetilde{\mathcal{O}}$ absorbs polylogarithmic factors in $T$ and $1/\delta$.
\end{theorem}

Conditions~\ref{asm:surrogate_risk}--\ref{asm:heb} are intentionally modular: in Section~\ref{sec:analysis} we verify them for O2PL using clipped importance-weighted losses, a self-normalized maximal inequality, and the H\"olderian error-bound condition.


\section{Regret analysis}
\label{sec:analysis}

We now instantiate the master theorem for O2PL by verifying its conditions in sequence. First, we verify surrogate excess-risk control via pessimismic policy learning (Condition~\ref{asm:surrogate_risk}). We introduce a new maximal inequality to obtain uniform confidence control for the surrogate objective. Second, we control the surrogate bias (Condition~\ref{asm:surrogate_bias}). Eventually, the variance bound (Condition~\ref{asm:heb}) is immediate from the H\"olderian error bound (Condition~\ref{asm:HEB}).

\subsection{Excess surrogate risk control via pessimistic policy learning}

For every $t \geq 1$, let $\widehat R_t$ be an $\mathcal{F}_t$-measurable map $\Pi \to \mathbb{R}$, an estimator of $R_{t,\alpha_t}$, with $(\alpha_t)_{t \geq 1}$ given. We suppose first that we can construct uniform-in-$\Pi$ confidence intervals with widths depending on the estimated variance $\hat \sigma_{t,\alpha}(\pi)$. 

\begin{condition}[Uniform confidence intervals]
\label{asm:unif_CI}
There exists a known $\Phi : (0,\infty) \times \mathbb{N} \times (0,\infty)\times (0,1) \to (0,\infty)$ and, for every $t \ge 1$, an $\mathcal{F}_t$-measurable $\widehat\sigma_{t,\alpha} : \Pi \to (0,\infty)$ such that with probability at least $1-\delta_t$, for all $\pi \in \Pi$,
\begin{align}
\label{eq:unif_CI}
\big| R_{t,\alpha_t}(\pi) - \widehat R_t(\pi) \big|
\le
\Phi(\widehat\sigma_{t,\alpha}(\pi),t, \alpha_t,\delta_t).
\end{align}
\end{condition}
We define the pessimistic policy at round $t+1$ as
\begin{equation}
 \label{eq:surro_ci}
\pi_{t+1}
\in
\arg\min_{\pi \in \Pi}
\widehat R_t(\pi)
+
\Phi(\widehat\sigma_{t,\alpha}(\pi),t, \alpha_t, \delta_t).
\end{equation}

\begin{condition}[Empirical to predictable variance]
\label{asm:emp_to_pred_var}
With probability at least $1-\delta_t$:
$$\Phi(\widehat \sigma_{t,\alpha}(\pi^\star), t, \alpha_t, \delta_t) \lesssim \bar \sigma_t^{1-b} / \sqrt{t} + r_{t,\alpha}^2$$.
\end{condition}
In the next subsection \ref{sec:maximal_inequality}, we verify the Conditions~\ref{asm:unif_CI}, \ref{asm:emp_to_pred_var}. 

\begin{proposition}\label{prop:pessimism_to_surrogate}
Suppose Conditions~\ref{asm:unif_CI} and~\ref{asm:emp_to_pred_var} hold. Then, with probability at least $1-2\delta_t$, Condition~\ref{asm:surrogate_risk} holds.
\end{proposition}

\begin{proof}
Fix $t\ge1$ and work on the event where \eqref{eq:unif_CI} holds uniformly over $\Pi$. By the pessimistic selection rule \eqref{eq:surro_ci},
\begin{align}
R_{t,\alpha_t}(\pi_{t+1})
&\le \widehat R_t(\pi_{t+1}) + \Phi(\widehat\sigma_{t,\alpha}(\pi_{t+1}),t, \alpha_t, \delta_t) \nonumber\\
&\le \widehat R_t(\pi^\star) + \Phi(\widehat\sigma_{t,\alpha}(\pi^\star),t,\alpha_t, \delta_t) \nonumber\\
&\le R_{t,\alpha_t}(\pi^\star) + 2\,\Phi(\widehat\sigma_{t,\alpha}(\pi^\star),t,\alpha_t,\delta_t),
\end{align}
hence
\[
R_{t,\alpha_t}(\pi_{t+1}) - R_{t,\alpha_t}(\pi^\star) \le 2\,\Phi(\widehat\sigma_{t,\alpha}(\pi^\star),t,\alpha_t,\delta_t).
\]
The claim follows from Condition~\ref{asm:emp_to_pred_var}.
\end{proof}

\subsection{A Maximal Inequality for Uniform inference on the surrogate policy risk}
\label{sec:maximal_inequality}
We seek a uniform control of the surrogate policy risk; to do so we verify Conditions~\ref{asm:unif_CI}, \ref{asm:emp_to_pred_var} by constructing the penalty
$\Phi$ in three steps. First, we state a novel maximal inequality for
martingale empirical processes localized by the \emph{realized}
quadratic variation, and specialize it to the clipped loss class; its
deviation bound is the functional $\widetilde H_t$
of~\eqref{eq:generic_deviation_radius}. Second, entropy transfer
(Proposition~\ref{prop:entropy_equivalence}) reduces $\widetilde H_t$ to
the static sup-norm geometry of $\log\Pi$, yielding the critical radius
$r_{\alpha,t}$. Third, peeling over variance shells self-normalizes the
bound, producing the $\pi$-dependent penalty $\Phi$
(Theorem~\ref{thm:peeling}) exhibited in Conditions~\ref{asm:unif_CI}, \ref{asm:emp_to_pred_var}.

\paragraph{Sequential bracketing entropy.}
For our maximal inequality, we use the bracketing
entropy in the realized norm as a complexity measure. For any $t\ge1$, index
set $\Theta$, and class of adapted processes
$\{\xi_{1:t}(\theta):\theta\in\Theta\}$ with each $\xi_s(\theta)$ being
$\mathcal F_s$-measurable, define the realized empirical norm
\[
    \widehat \rho_t(\xi_{1:t}(\theta))
    := \Bigl( \tfrac{1}{t} \textstyle\sum_{s=1}^t \xi_s(\theta)^2 \Bigr)^{1/2}.
\]
An $\epsilon$-sequential bracketing is a family
$\{(\Lambda^j_{1:t},\Gamma^j_{1:t})\}_{j=1}^N$ of $\mathcal F_s$-measurable
brackets such that every $\xi_{1:t}(\theta)$ is trapped
($\Lambda^j_s\le\xi_s(\theta)\le\Gamma^j_s$ for all $s$) by some bracket
of realized width $\widehat\rho_t(\Gamma^j-\Lambda^j)\le\epsilon$; we
write $N_{[\,]}(\epsilon,\Theta)$ for the smallest such $N$ and
\[
  J_{[\,]}(\sigma,\Theta)
  := \int_0^\sigma \sqrt{\log N_{[\,]}(\epsilon,\Theta)}\,d\epsilon
\]
for the associated entropy integral.

\paragraph{The maximal inequality.} We state below our novel uniform deviation bound on which the confidence radii in our analysis are built.  

\begin{theorem}[Maximal inequality under realized-variation localization]
\label{thm:selfnorm-chaining-body}
Let $\Theta$ index $\mathcal F_s$-measurable processes with
$|\xi_s(\theta)|\le\alpha$ a.s., and set
\[
  M_t(\theta):=\tfrac1t\textstyle\sum_{s=1}^t\bigl(\xi_s(\theta)-\E[\xi_s(\theta)\mid\mathcal F_{s-1}]\bigr).
\]
There is a universal $C>0$ such that, for every $\sigma>0$ and
$\delta\in(0,1]$, with probability at least $1-\delta$,
\[
  \sup_{\theta\in\Theta}
    \mathbf 1_{\{\widehat\rho_t(\xi_{1:t}(\theta))\le\sigma\}}\,
    M_t(\theta)
  \;\le\;
  C\,\widetilde H_t(\sigma,\alpha,\delta),
\]
where
\begin{align}
    \widetilde H_t(\sigma,\alpha,\delta)
    &:=
    \frac{1}{\sqrt t}\,J_{[\,]}(\sigma,\Theta)
    + \frac{\alpha}{t}\,\log N_{[\,]}(\sigma,\Theta)
    \nonumber\\
    &\quad
    + \sigma\sqrt{\tfrac{\log(e/\delta)}{t}}
    + \alpha\,\tfrac{\log(e/\delta)}{t}.
    \label{eq:generic_deviation_radius}
\end{align}
\end{theorem}

The proof is deferred to Appendix~\ref{app:maximal_inequality}. The key novelty is that our maximal inequality localizes the martingale empirical process by the \emph{realized} quadratic variation, rather than by the usual predictable variation. This realized, hence observable, quadratic control is what makes it possible to chain over the policy class while retaining a data-dependent variance proxy under adaptive sampling. The bound has the same Bernstein-type structure as the scalar inequality in Proposition~\ref{prop:self-normalized-crossing}: the first two terms of $\widetilde H_t$ are the localized complexity contribution, while the last two are the standard Bernstein corrections, matching the classical scalar decomposition \citep{freedman1975, delaPeña2009}.

\paragraph{Application to the clipped loss class.}
We apply Theorem~\ref{thm:selfnorm-chaining-body} to the clipped
importance-weighted loss $\ell_{s,\alpha}(\pi)$, an adapted process
bounded with an envelope $\alpha$ and indexed by $\pi\in\Pi$, whose
centered average is the deviation $\widehat R_t(\pi)-R_{t,\alpha}(\pi)$
of the empirical clipped objective~\eqref{eq:empirical_risk} from its
predictable target. Since the inequality localizes by the realized norm,
the class it must be applied to is the one restricted to a variance
shell,
\[
    \mathcal L^\alpha_{1:t}(\sigma)
    := \bigl\{ \ell_{1:t,\alpha}(\pi) : \widehat \rho_t( \ell_{1:t,\alpha}(\pi))\le \sigma,\ \pi\in\Pi \bigr\},
\]
so the entropy entering $\widetilde H_t(\sigma,\alpha,\delta)$
in~\eqref{eq:generic_deviation_radius} is that of this localized class,
$N_{[\,]}(\epsilon,\mathcal L^\alpha_{1:t}(\sigma))$. Controlling
$\widetilde H_t$ therefore reduces to bounding this localized entropy,
which the next proposition transfers to the sup-norm geometry of
$\log\Pi$; peeling over the shells $\sigma$ then yields the
$\pi$-dependent penalty $\Phi$.

\paragraph{Entropy transfer to the log-policy class.}
The next proposition shows that the
localized complexity of $\mathcal L^\alpha_{1:t}(\sigma)$ is governed by the sup-norm geometry of $\log\Pi$.

\begin{proposition}[Entropy transfer on the log scale]
\label{prop:entropy_equivalence}
Assume $Y_s\in[-1,0]$ a.s.\ and $\alpha\ge 1$. For every $t\ge 1$,
$\sigma>0$, $\epsilon>0$, and a.e.\ realization,
\[
N_{[\,]}(\epsilon,\mathcal L^\alpha_{1:t}(\sigma))
\le
N\!\Bigl(\tfrac{c\epsilon}{1+\sigma},\log\Pi,\|\cdot\|_\infty\Bigr)
\]
for a universal constant $c>0$.
\end{proposition}

The mechanism is simple: ratios satisfy
$\tfrac{\pi}{\pi_t}(a\mid x)=\exp(\log\pi(a\mid x)-\log\pi_t(a\mid x))$,
so sup-norm control of $\log\pi$ gives multiplicative control of the
ratios; since $w\mapsto(\min(w,\alpha)-1)y$ is monotone and Lipschitz
for $y\in[-1,0]$, brackets transfer directly (proof in
Appendix~\ref{app:transfer_inequality}).

\paragraph{Localized complexity and critical radius.}
Combining Proposition~\ref{prop:entropy_equivalence} with
Condition~\ref{asm:entropy_logPi_scale} gives, in the small-radius
regime, $\log N_{[\,]}(\epsilon,\mathcal L^\alpha_{1:t}(\sigma))\lesssim
\log(e/\epsilon)$ for $p=0$ and $\lesssim\epsilon^{-p}$ for
$p\in(0,2)$, up to constants depending on $1+\sigma$. Substituting
into~\eqref{eq:generic_deviation_radius},
\begin{equation}
    \widetilde H_t(\sigma,\alpha_t,\delta)
    \lesssim
    \mathcal H_t(\sigma,\alpha)
    + \sigma\sqrt{\tfrac{\log(e/\delta)}{t}}
    + \alpha_t\tfrac{\log(e/\delta)}{t},
    \label{eq:Ht_upper_bound}
\end{equation}
where
\[
\mathcal H_t(\sigma,\alpha)
:=
\begin{cases}
\dfrac{\alpha_t}{t}\log(e/\sigma)
+\dfrac{\sigma}{\sqrt t}\sqrt{\log(e/\sigma)},
& p=0,\\[1em]
\dfrac{\alpha_t}{t}\sigma^{-p}
+\dfrac{\sigma^{1-p/2}}{\sqrt t},
& p\in(0,2).
\end{cases}
\]
Solving $\mathcal H_t(r,\alpha)=r^2$ from
\eqref{eq:critical_radius_abstract} gives
$r_{\alpha,t}\asymp(\alpha_t/t)^{1/(2+p)}$, with the same scaling up to
logarithmic factors when $p=0$.

Appendix~\ref{sec:peeling} turns Equation \eqref{eq:Ht_upper_bound} into a
self-normalized maximal inequality by peeling over the shells of
$\hat\rho_t(\ell_{1:t,\alpha}(\pi))$. Since
$\widehat\sigma_{t,\alpha_t}(\pi)$ upper-bounds
$\hat\rho_t(\ell_{1:t,\alpha}(\pi))$ which results in the following theorem:

\begin{theorem}[Self-normalized confidence bound]
\label{thm:peeling}
Under Condition~\ref{asm:entropy_logPi_scale} with $p\in[0,2)$, there is
a universal $C>0$ such that with probability at least $1-\delta$, every
$\pi\in\Pi$ satisfies
\begin{align*}
  \bigl|\widehat R_t(\pi)-R_{t,\alpha_t}(\pi)\bigr|
  &\lesssim
    r_{\alpha,t}^2
    + \frac{\widehat \sigma_{t,\alpha_t}(\pi)^{\,1-p/2}}{\sqrt t}
  \\
  &\qquad
    + \widehat \sigma_{t,\alpha_t}(\pi)\sqrt{\tfrac{\log(et/\delta)}{t}}
    + \alpha_t\tfrac{\log(et/\delta)}{t}
\end{align*}
with the second term interpreted up to a logarithmic factor when $p=0$.
\end{theorem}

This bound is precisely the penalty $\Phi$ of
Condition~\ref{asm:unif_CI}: $r_{\alpha,t}^2$ is the critical-radius
floor, $\widehat\sigma_{t,\alpha_t}(\pi)^{1-p/2}/\sqrt t$ is the
localization gain from small empirical variance, and the rest are
standard confidence corrections. Combining Theorem~\ref{thm:peeling}
with Proposition~\ref{prop:pessimism_to_surrogate} and the concentration
step transferring $\widehat\sigma_{t,\alpha_t}(\pi^\star)$ to
$\bar\sigma_t$ yields:

\begin{proposition}[Verification of Condition~\ref{asm:emp_to_pred_var}]
\label{prop:surrogate_risk_verification}
Under Conditions~\ref{asm:entropy_logPi_scale} and~\ref{asm:HEB}, with
probability at least $1-2\delta_t$,
$R_{t,\alpha_t}(\pi_{t+1})-R_{t,\alpha_t}(\pi^\star) \lesssim
\bar\sigma_t^{\,1-p/2}/\sqrt t + r_{\alpha,t}^2$. In particular,
Condition~\ref{asm:surrogate_risk} then holds with $b=p/2$.
\end{proposition}

\begin{remark}[Why the variance estimator uses clipped weights]
The pessimistic penalty uses the variance estimator
$\widehat\sigma_{t,\alpha_t}(\pi)$ of the clipped surrogate losses. This
is natural, since the confidence analysis is carried out for the clipped
class, and convenient: clipping ensures bounded increments for the
squared loss process, allowing the empirical proxy at $\pi^\star$ to be
compared to the predictable variance via Freedman's inequality. An
unclipped analogue would require an extra boundedness or tail assumption
on $\pi^\star/\pi_t$, essentially an optimal-action overlap
$\pi_t(a^\star(x)\mid x)>\eta>0$.
\end{remark}

\subsection{Surrogate bias control}

The final step is to control the clipping approximation error. Standard analyses bound it by the clipping level alone, a deterministic $O(1/\alpha)$ that does not adapt to the difficulty of the problem. Our observation is that clipping bias is instead controlled by the \emph{variance of the importance weights}: large ratios---the only ones clipping affects---are rare when $\sigma_s^2(\pi)$ is small, so the bias inherits the low-noise structure.

\begin{lemma}[Clipping bias]\label{lem:clipping_bias}
For any $\pi\in\Pi$ and $\alpha>1$,
\begin{equation}
R_t(\pi)
\le
R_{t,\alpha}(\pi)
\le
R_t(\pi)
+
\frac{1}{t(\alpha-1)}\sum_{s=1}^t \sigma_s^2(\pi).
\label{eq:clipping_bias_bound}
\end{equation}
\end{lemma}

The proof (Appendix~\ref{app:clipping_bias}) simply uses Chebyshev's inequality and the astute observation that clipping can be decomposed as: $$ \min(\alpha,\frac{\pi}{\pi_t})= \frac{\pi}{\pi_t} -\big(\frac{\pi}{\pi_t}-\alpha\big)_+$$ This is how the H\"olderian bound propagates into the bias: near-optimal policies have small $\sigma_s^2(\pi^\star)$, hence small clipping bias, permitting slower growth of $\alpha_t$ and a tighter critical radius. Consequently Condition~\ref{asm:surrogate_bias} holds, since for any $\pi$ and $\alpha>1$,
\begin{align}
&\big(R_t(\pi)-R_t(\pi^\star)\big)
-
\big(R_{t,\alpha}(\pi)-R_{t,\alpha}(\pi^\star)\big)
\nonumber\\
&\qquad\le
R_{t,\alpha}(\pi^\star)-R_t(\pi^\star)
\le
\frac{1}{\alpha-1}\bar \sigma_t^2,
\label{eq:surrogate_bias_excess_risk_bandit}
\end{align}
and $R_t(\pi)-R_t(\pi^\star)=R(\pi)-R(\pi^\star)$ gives the required control.

\section{Numerical investigation}
\label{sec:numerics}

We report a compact numerical study organized around four questions aligned with the theory:
(i) does pessimistic learning improve performance, and in particular does variance penalization help under dependent data?
(ii) how does our agnostic policy-based method compare to other agnostic policy-based methods?
(iii) how does it compare to value-based methods under misspecification?
(iv) does regret scaling track the H\"olderian exponent?
Additional experiments, implementation details, and comparisons are deferred to Appendix~\ref{app:additional_numerical}.

\begin{figure*}[!t]
\centering
\includegraphics[
  width=0.9\textwidth,
  height=0.34\textheight,
  keepaspectratio
]{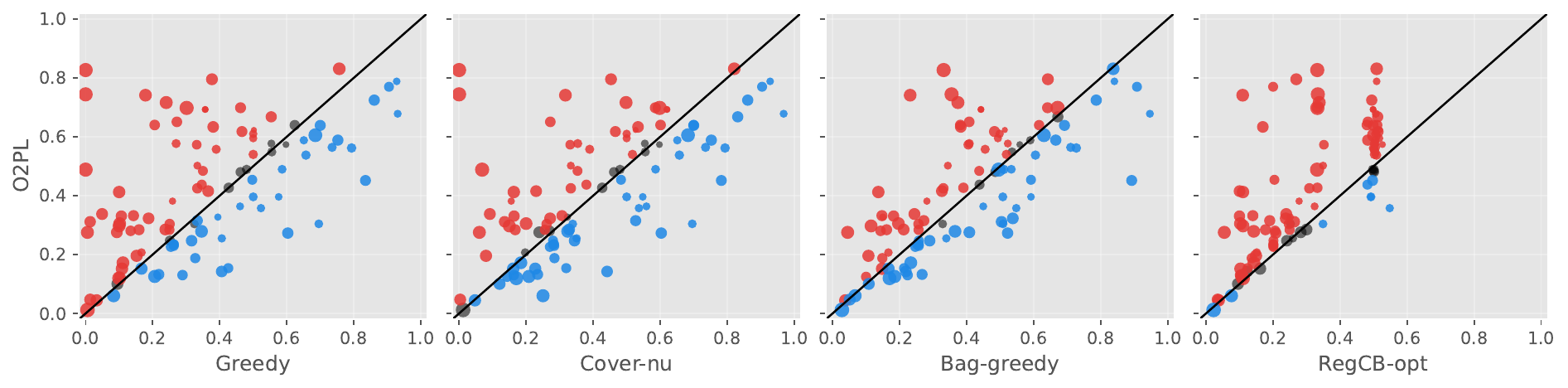}
\caption{OpenML benchmark: pairwise normalized performance. Red points denote significant O2PL wins, blue points significant O2PL losses, and black points non-significant comparisons, according to a reward-based one-sided $Z$-test. }
\label{fig:openml_bakeoff}
\end{figure*}

\subsection{Variance penalization under temporal dependence}

\begin{figure}[H]
  \centering
  \includegraphics[width=0.7\linewidth]{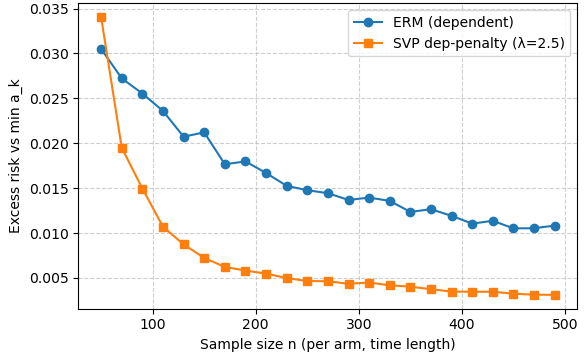}
  \caption{ERM vs.\ variance-penalized selection under dependent data.}
  \label{fig:svp_experiment}
\end{figure}

We first isolate the role of pessimism by comparing ERM to a variance-penalized variant under temporally dependent observations. Our self-normalized maximal inequality is designed for this setting: it provides confidence control in terms of an observable realized variance proxy, and therefore motivates a dependent-data analogue of the sample variance penalization (SVP) principle of \citet{maurer2009empirical}. We adapt their classical SVP-versus-ERM experiment by replacing iid observations with coordinate-wise two-state Markov chains. For each coordinate $k$, we set $X_{t,k}=a_k+b_k S_{t,k}$ with $S_{t,k}\in\{-1,+1\}$ a two-state Markov chain of persistence $p_k$, so lag-one correlation $\chi_k=2p_k-1$. As in the iid setting, ERM ($\lambda=0$) can over-select high-variance arms that look favorable by noise; dependence amplifies this because effective sample sizes shrink. We compare ERM to a variance-penalized score: $$\widehat{\mu}_k+\lambda\sqrt{\widehat v_k/n_{\mathrm{eff},k}} \quad \text{with }n_{\mathrm{eff},k}=n(1-\widehat\chi_k)/(1+\widehat\chi_k)$$. Figure~\ref{fig:svp_experiment} shows a clear gap in favor of variance penalization.

\subsection{Comparison to other agnostic methods}

We also compare O2PL against other agnostic policy-learning baselines on the OpenML benchmark suite \cite{bischl2021openmlbenchmarkingsuites} where we do the standard multiclass-to-contextual-bandit conversion of \citet{bietti2021contextualbanditbakeoff}. 
O2PL tends to outperform competing methods as environment complexity increases; Figure~\ref{fig:openml_bakeoff} reports pairwise normalized performance. Additional details are in Appendix~\ref{app:additional_numerical}.

\subsection{Misspecification: the wheel bandit}

\begin{figure}[H]
  \centering
  \includegraphics[width=0.72\linewidth]{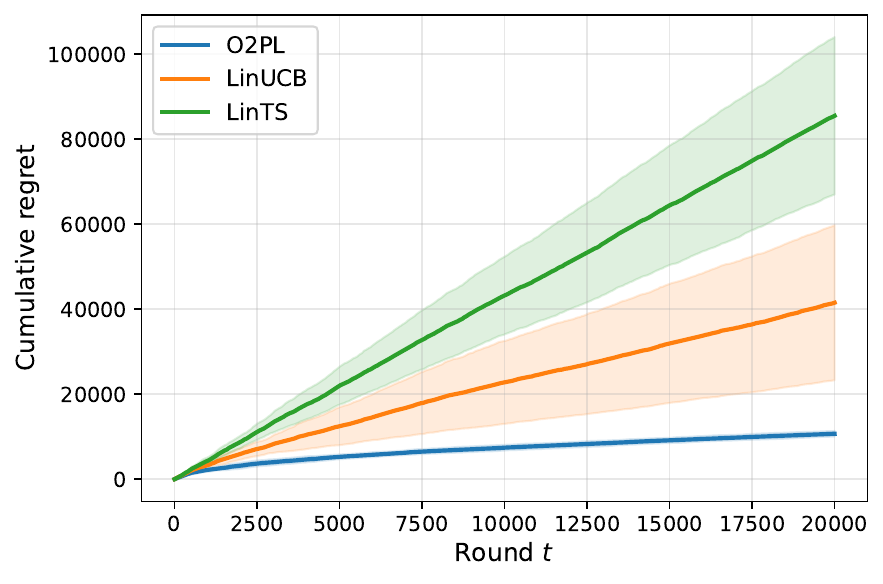}
  \caption{Cumulative regret on the wheel bandit  \\with $\delta=0.95$ and $(\mu_1,\mu_2,\mu_3)=(1.2,1,50)$.}
  \label{fig:misspecification}
\end{figure}

We compare our agnostic policy based method to value based methods in a setting where we mispecify the value function on purpose. We use the wheel bandit environment \cite{riquelme2018deepbayesianbanditsshowdown}: Contexts are uniform on the unit disk in $\mathbb{R}^2$ with five actions. Action $0$ is a \emph{safe} arm with constant reward $\mu_1$; the other four depend on context. Inside the ball $\|x\|\leq \delta$ all four are suboptimal with mean $\mu_2 < \mu_1$; outside, exactly one becomes optimal per quadrant with reward $\mu_3 \gg \mu_1$. The optimal policy is inherently nonlinear---safe arm near the center, quadrant-dependent outside---so when $\delta$ is large the high-reward region is rare and linear methods like LinUCB/LinTS can be misled into the safe arm. Because O2PL learns policies directly from bandit feedback rather than fitting a globally linear reward model, it is more robust; Figure~\ref{fig:misspecification} shows it substantially outperforms LinUCB and LinTS.

\subsection{Regret scaling with the H\"olderian exponent}

We examine how regret scales with the H\"olderian exponent in a controlled synthetic environment.

\begin{figure}[H]
  \centering
  \includegraphics[width=0.72\linewidth]{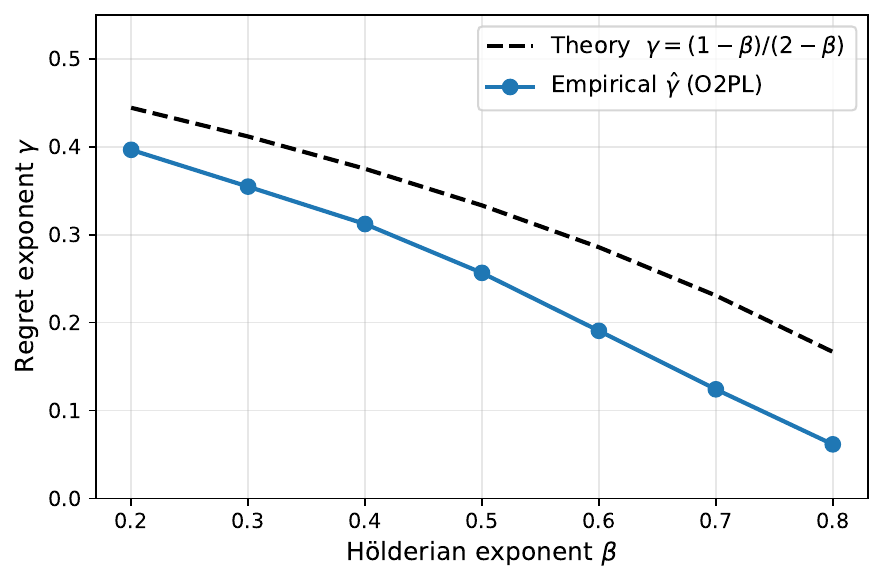}
  \caption{Empirical versus theoretical regret exponent as $\beta$ varies. Larger $\beta$ leads to slower regret growth, consistent with theory.}
  \label{fig:beta_exponent}
\end{figure}

Let $x_t=s_t z_t$ with $s_t\in\{-1,1\}$ Rademacher and $z_t\sim\mathrm{Beta}(\nu,1)$, and choose mean rewards so $\Delta(x)\propto |x|$; then $\Pr(\Delta\le u)\asymp u^\nu$, yielding $\beta=\nu/(\nu+1)$. Varying $\nu$ tunes $\beta$. Figure~\ref{fig:beta_exponent} shows that as $\beta$ increases, observed regret growth slows, consistent with theory. Moreover, the empirical regret exponent variation matches that of the theoretical regret.

\section{Conclusion}

We studied agnostic policy-class contextual bandits and proved fast best-in-class regret for a fully online algorithm with every-round updates. The algorithmic ingredient is pessimistic policy improvement with variance-aware penalties; the main probabilistic ingredient is a new sequential self-normalized maximal inequality for martingale empirical processes, localized by realized quadratic variation. This inequality yields uniform variance-adaptive confidence bounds under adaptive sampling and is of separate interest beyond O2PL. Our regret analysis is organized through a master theorem: any algorithm satisfying surrogate-risk, bias, and variance conditions inherits fast rates, and O2PL verifies these conditions through clipping, pessimism, and the H"olderian error bound. The remaining open issue is computational: O2PL calls a variance-penalized policy-learning oracle at every round, and this objective is generally nonconvex. Understanding when practical approximations, such as majorization--minimization for variance-penalized IPW objectives \citep{swaminathan2012}, preserve the fast-rate guarantee remains an important direction.

\bibliography{references}

\newpage
\onecolumn
\appendix
\section*{Appendix}

This appendix is organized as follows:
\begin{itemize}[nosep, label={--}]
    \item Appendix~\ref{app:realizability}: proof that realizability
      and Tsybakov margin imply the H\"olderian error bound
      (Lemma~\ref{lem:margin_implies_heb}).
    \item Appendix~\ref{app:maximal_inequality}: the core probability
      tools---Bernstein inequality with realized variation
      (Proposition~\ref{prop:self-normalized-crossing}), sequential
      chaining, and the conditional expectation bound with
      realized-variation localization
      (Theorem~\ref{thm:selfnorm-chaining}).
    \item Appendix~\ref{app:complexity_control}: complexity control
      and uniform confidence---entropy transfer on the log scale
      (Proposition~\ref{prop:entropy_equivalence}), bounding the
      complexity measure via the critical radius, self-normalized
      confidence bound via peeling (Theorem~\ref{thm:peeling}), and
      verification of Condition~\ref{asm:emp_to_pred_var}.
    \item Appendix~\ref{app:regret_analysis}: regret analysis---clipping
      bias control (Lemma~\ref{lem:clipping_bias}) and proof of the
      master theorem (Theorem~\ref{thm:regret}).
    \item Appendix~\ref{app:additional_numerical}: implementation
      details, additional experiments, and supplementary results.
\end{itemize}
\section{Sufficient condition for the Holderian error bound }
\label{app:realizability}

Lemma~\ref{lem:margin_implies_heb} provides one concrete sufficient
condition for the H\"olderian error bound, but
Condition~\ref{asm:HEB} itself is a single scalar inequality on the
policy class that does not explicitly require realizability, a margin
gap, or optimal-action overlap. It captures only their joint
consequence: that importance-weight variance vanishes polynomially
with excess risk. This makes it amenable to verification in settings
beyond the realizable-margin framework of
Lemma~\ref{lem:margin_implies_heb}

\begin{proof}[Proof of Lemma~\ref{lem:margin_implies_heb}]
Applying the margin condition at $u=0$ yields $\Pr(\Delta(X)=0)=0$, so Bayes-optimal actions are unique almost surely. Therefore $a^\star(X)$ is well-defined almost surely.

Because $\pi^\star$ attains Bayes risk and each non-Bayes action has strictly larger conditional mean loss almost surely, necessarily
\[
\pi^\star(a^\star(X)\mid X)=1\qquad\text{a.s.}
\]
Fix any $\pi\in\Pi$ and define
\[
q_\pi(X):=1-\pi(a^\star(X)\mid X)\in[0,1].
\]
Using $\pi^\star(\cdot\mid X)=\delta_{a^\star(X)}$, we have pointwise
\[
\chi^2(\pi^\star(\cdot\mid X)\|\pi(\cdot\mid X))
=\frac{1}{\pi(a^\star(X)\mid X)}-1
=\frac{q_\pi(X)}{\pi(a^\star(X)\mid X)}
\le \frac{q_\pi(X)}{c_0},
\]
hence, taking expectation over $X$,
\[
\Var\!\left(\frac{\pi^\star}{\pi}(A\mid X)\right)
=\E\!\left[\chi^2(\pi^\star(\cdot\mid X)\|\pi(\cdot\mid X))\right]
\le \frac{1}{c_0}\E[q_\pi(X)].
\]

It remains to bound $\E[q_\pi(X)]$ by excess risk. Let $(X,A)\sim P_X\times\pi(\cdot\mid X)$. Then
\[
R(\pi)-R(\pi^\star)
\;=\;
\E\!\left[\sum_{a\in\mathcal A}\pi(a\mid X)\big(\mu(X,a)-\mu^\star(X)\big)\right]
\;\ge\;
\E\!\left[q_\pi(X)\Delta(X)\right]
 =: r.
\]
Now fix any $u>0$ and decompose:
\[
\E[q_\pi(X)]
\le \E[\mathbf 1\{\Delta(X)\le u\}]+\E[q_\pi(X)\mathbf 1\{\Delta(X)>u\}]
\le \Pr(\Delta(X)\le u)+\frac{\E[q_\pi(X)\Delta(X)]}{u}
\le C_{\mathrm m}u^\kappa+\frac{r}{u}.
\]
The right-hand side is minimized at
\[
u^\star=\left(\frac{r}{\kappa C_{\mathrm m}}\right)^{\frac{1}{\kappa+1}},
\]
which gives
\[
\E[q_\pi(X)]
\le
\left(\kappa^{\frac{1}{\kappa+1}}+\kappa^{-\frac{\kappa}{\kappa+1}}\right)
C_{\mathrm m}^{\frac{1}{\kappa+1}}
r^{\frac{\kappa}{\kappa+1}}
=:C_{\kappa,C_{\mathrm m}}\,r^{\frac{\kappa}{\kappa+1}}.
\]
Since $r\le R(\pi)-R(\pi^\star)$, we conclude
\[
\Var\!\left(\frac{\pi^\star}{\pi}(A\mid X)\right)
\le
\frac{C_{\kappa,C_{\mathrm m}}}{c_0}\,(R(\pi)-R(\pi^\star))^{\frac{\kappa}{\kappa+1}},
\]
which is the claim.
\end{proof}


\section{Self-normalized maximal inequality}
\label{app:maximal_inequality}

The maximal inequality of Theorem~\ref{thm:selfnorm-chaining}
controls the supremum of a martingale array localized by the
\emph{realized} quadratic variation
$\hat\rho_n(X^\theta)^2=n^{-1}\sum_{i=1}^n(X_i^\theta)^2$.  This
quantity is $\mathcal F_n$-measurable but not predictable: it depends
on the full sample path and cannot be expressed as a sum of
conditional variances.  The classical chaining argument
of~\cite{van_Handel_2010} localizes instead by the \emph{predictable}
variation $\sum_{i=1}^n\E[(X_i^\theta)^2\mid\mathcal F_{i-1}]$,
which is a predictable process and amenable to standard Freedman-type
inequalities.

To work with the realized variation, we require a concentration
inequality in which the quadratic control is itself
non-predictable.  Proposition~\ref{prop:self-normalized-crossing}
below provides such a bound: it is a Bernstein-type concentration bound 
 where the variance term is the observed sum of squares
$S_j=\sum_{t=1}^j X_t^2$ rather than its conditional expectation.
Once this ingredient is in place, the maximal inequality and chaining
argument proceed along the lines of~\cite{van_Handel_2010}, with the
predictable variation replaced throughout by the realized variation.

\subsection{A Freedman-type inequality with realized quadratic control}

\begin{proposition}[Bernstein inequality with the realizable variation]
\label{prop:self-normalized-crossing}
Let $(\mathcal F_t)_{t\ge 0}$ be a filtration and let
$(X_t)_{t\ge 1}$ be a sequence of real-valued random variables such
that $X_t$ is $\mathcal F_t$-measurable and $|X_t|\le B$ a.s.\ for
all $t\ge 1$.  Define
\[
  M_j := \sum_{t=1}^j
    \bigl(X_t-\E[X_t\mid\mathcal F_{t-1}]\bigr),
  \qquad
  S_j := \sum_{t=1}^j X_t^2,
  \qquad j\ge 1,
\]
with $M_0=S_0=0$.  Then for all $u,v>0$,
\[
  \Pr\!\bigl(\exists\, j\ge 1:\;
    M_j\ge u,\; S_j\le v\bigr)
  \;\le\;
  \exp\!\Bigl(-\frac{u^2}{4(v+Bu)}\Bigr).
\]
\end{proposition}

\begin{proof}
Write $D_t=X_t-\E[X_t\mid\mathcal F_{t-1}]$ for the martingale
difference.

\medskip\noindent
\textbf{Step~1: a scalar inequality.}
For every $x\ge -1/2$,
\begin{equation}\label{eq:key-scalar-ineq}
  e^{x-x^2}\le 1+x.
\end{equation}
Indeed, $g(x):=\log(1+x)-x+x^2$ satisfies
$g'(x)=x(1+2x)/(1+x)$, which is nonpositive on $[-1/2,0]$ and
nonnegative on $[0,\infty)$.  Since $g(0)=0$, we have $g(x)\ge 0$
for all $x\ge -1/2$, and exponentiating
gives~\eqref{eq:key-scalar-ineq}.

\medskip\noindent
\textbf{Step~2: one-step bound.}
Fix $\lambda\in(0,1/(2B)]$.  Since $|\lambda X_t|\le 1/2$,
applying~\eqref{eq:key-scalar-ineq} with $x=\lambda X_t$ and
taking conditional expectations yields
\[
  \E\bigl[e^{\lambda X_t-\lambda^2 X_t^2}
    \mid\mathcal F_{t-1}\bigr]
  \;\le\; 1+\lambda\,\E[X_t\mid\mathcal F_{t-1}].
\]
Multiplying both sides by
$e^{-\lambda\,\E[X_t\mid\mathcal F_{t-1}]}$ and using the
elementary inequality $e^{-y}(1+y)\le 1$ for all $y\in\mathbb R$,
we obtain
\begin{equation}\label{eq:one-step}
  \E\bigl[e^{\lambda D_t-\lambda^2 X_t^2}
    \mid\mathcal F_{t-1}\bigr]
  \;\le\; 1.
\end{equation}

\medskip\noindent
\textbf{Step~3: supermartingale and stopping.}
Define
$\mathcal E_j(\lambda):=\exp(\lambda M_j-\lambda^2 S_j)$.
The multiplicative decomposition
$\mathcal E_j=\mathcal E_{j-1}\cdot
  \exp(\lambda D_j-\lambda^2 X_j^2)$
together with~\eqref{eq:one-step} shows that
$(\mathcal E_j(\lambda))_{j\ge 0}$ is a positive supermartingale
with $\mathcal E_0=1$.

Let $\tau=\inf\{j\ge 1:M_j\ge u,\;S_j\le v\}$.
On $\{\tau\le m\}$ we have
$\mathcal E_\tau(\lambda)\ge\exp(\lambda u-\lambda^2 v)$, so the
optional stopping theorem applied to $\tau\wedge m$ gives
$\Pr(\tau\le m)\le\exp(-\lambda u+\lambda^2 v)$ for every $m\ge 1$.
Sending $m\to\infty$ by continuity from below,
\begin{equation}\label{eq:lambda-bound}
  \Pr\!\bigl(\exists\, j\ge 1:M_j\ge u,\;S_j\le v\bigr)
  \;\le\;
  \exp(-\lambda u+\lambda^2 v).
\end{equation}

\medskip\noindent
\textbf{Step~4: optimization over $\lambda$.}
Minimizing $f(\lambda)=-\lambda u+\lambda^2 v$ over
$\lambda\in(0,1/(2B)]$ gives two cases.  If the unconstrained
minimizer $\lambda_*=u/(2v)$ is admissible (i.e.\ $u\le v/B$),
then $f(\lambda_*)=-u^2/(4v)$.  Otherwise, evaluating at the
boundary $\lambda=1/(2B)$ and using $v<Bu$ gives
$f(1/(2B))<-u/(4B)$.  In both cases,
\[
  f(\lambda)
  \;\le\;
  -\min\Bigl\{\frac{u^2}{4v},\;\frac{u}{4B}\Bigr\}
  \;\le\;
  -\frac{u^2}{4(v+Bu)},
\]
where the last inequality follows from
$u^2/(4v)\ge u^2/(4(v+Bu))$ and $u/(4B)=u^2/(4Bu)\ge u^2/(4(v+Bu))$.
Substituting into~\eqref{eq:lambda-bound} completes the proof.
\end{proof}
\subsection*{B.2. Maximal inequality for a finite family}

\paragraph{Notation.}
For an event $A\in\mathcal F$ with $\Pr[A]>0$ and an integrable random
variable $Z$, we write
\[
  \E^A[Z]\;:=\;\frac{\E[Z\,\mathbf 1_A]}{\Pr[A]}\;=\;\E[Z\mid A].
\]
We keep the event $A$ explicit as a free parameter: the results below
hold \emph{simultaneously for every choice of $A$}, and this is what
lets the high-probability form of Theorem~\ref{thm:selfnorm-chaining}
be recovered by a direct substitution, with no separate concentration
argument or union bound needed.

\begin{lemma}[Maximal inequality for a finite family]
\label{lem:self-normalized-finite}
Let $N\ge 1$, and let $X_1,\dots,X_N$ be real-valued random variables.
Suppose there exist constants $v>0$ and $B>0$ such that
\[
  \Pr\bigl(|X_i|\ge x\bigr)
  \;\le\;
  \exp\!\Bigl(-\frac{x^2}{4(v+Bx)}\Bigr)
  \qquad\text{for all } x>0 \text{ and all } 1\le i\le N.
\]
Then for every event $A\in\mathcal F$ with $\Pr[A]>0$,
\[
  \E^A\Bigl[\max_{1\le i\le N}|X_i|\Bigr]
  \;\le\;
  4\sqrt{v\log\!\Bigl(1+\frac{N}{\Pr[A]}\Bigr)}
  \;+\;16B\log\!\Bigl(1+\frac{N}{\Pr[A]}\Bigr).
\]
\end{lemma}

\begin{corollary}[Maximal inequality for a finite family of martingales]
\label{cor:self-normalized-finite}
Let $n\ge 1$, and let $(X_i^h)_{1\le i\le n}$, for $h=1,\dots,N$, be
$\mathcal F_i$-measurable arrays with $\|X_i^h\|_\infty\le B$. Then for
any event $A\in\mathcal F$ with $\Pr[A]>0$ and any $R>0$,
\[
  \E^A\Bigl[
    \max_{1\le h\le N}\;
    \mathbf 1_{\{\sum_{i=1}^n(X_i^h)^2\le R\}}\,
    \max_{1\le j\le n}\sum_{i=1}^j
      \bigl(X_i^h-\E[X_i^h\mid\mathcal F_{i-1}]\bigr)
  \Bigr]
  \;\le\;
  4\sqrt{R\log\!\Bigl(1+\frac{N}{\Pr[A]}\Bigr)}
  \;+\;16B\log\!\Bigl(1+\frac{N}{\Pr[A]}\Bigr).
\]
\end{corollary}

\bigskip
\subsection*{B.3. Chaining under realized-variation localization}

\begin{theorem}[Uniform deviation bound under realized-variation localization]
\label{thm:selfnorm-chaining}
Fix $n\ge 1$. Let $\Theta$ be a class of $\mathcal F_i$-measurable
arrays $X^\theta=(X_i^\theta)_{1\le i\le n}$ with $|X_i^\theta|\le B$
a.s.\ for all $i,\theta$. Define
\[
  M_n^\theta
  := \frac{1}{n}\sum_{i=1}^n
    \bigl(X_i^\theta-\E[X_i^\theta\mid\mathcal F_{i-1}]\bigr),
  \qquad
  \hat\rho_n(X^\theta)
  := \Bigl(\frac{1}{n}\sum_{i=1}^n(X_i^\theta)^2\Bigr)^{\!1/2}.
\]
For $\delta>0$, let $N(\delta,\Theta,\hat\rho_n)$ denote the smallest
number of predictable brackets $\{(\ell_p,u_p)\}_{p=1}^N$ such that
every $\theta\in\Theta$ is captured by some bracket with
$\ell_{p,i}\le X_i^\theta\le u_{p,i}$ and
$\hat\rho_n(u_p-\ell_p)\le\delta$. Define
\[
  H(R,B,\Theta,\hat\rho_n)
  :=
  \frac{B}{n}\log N(\sqrt R,\Theta,\hat\rho_n)
  +
  \frac{1}{\sqrt n}
  \int_0^{\sqrt R}
    \sqrt{\log N(u,\Theta,\hat\rho_n)}\,du.
\]
Then there exists a universal constant $C>0$ such that, for every
$R>0$ and every $\delta\in(0,1]$, with probability at least $1-\delta$,
\[
  \sup_{\theta\in\Theta}
    \mathbf 1_{\{\hat\rho_n(X^\theta)\le\sqrt R\}}\,M_n^\theta
  \;\le\;
  C\Bigg(
    H(R,B,\Theta,\hat\rho_n)
    +
    \sqrt{\frac{R}{n}\log\frac{2}{\delta}}
    +
    \frac{B}{n}\log\frac{2}{\delta}
  \Bigg).
\]
\end{theorem}

\begin{proof}
Before proceeding to the proof, let us define the function
\[
  \Phi(x)\;:=\;C\Bigl(H(R,B,\Theta,\hat\rho_n)+\sqrt{\tfrac{R}{n}\,x}
    +\tfrac{B}{n}\,x\Bigr), \qquad x\ge 0.
\]
We claim that in order to prove the Theorem, it actually suffices to
prove the estimate
\[
  \E^A\Bigl[\sup_{\theta\in\Theta}
    \mathbf 1_{\{\hat\rho_n(X^\theta)\le\sqrt R\}}\,M_n^\theta\Bigr]
  \;\le\;
  \Phi\Bigl(\log\frac{2}{\Pr[A]}\Bigr)
  \qquad\text{for every event }A\text{ with }\Pr[A]>0.
  \tag{$\ast$}
\]
Indeed, if this is the case, then, writing
$Z:=\sup_{\theta\in\Theta}\mathbf 1_{\{\hat\rho_n(X^\theta)\le\sqrt
R\}}M_n^\theta$, fixing $\delta\in(0,1]$ and choosing
$A:=\{Z\ge\Phi(\log(2/\delta))\}$ (the case $\Pr[A]=0$ being trivial),
we may estimate
\[
  \Phi\Bigl(\log\frac2\delta\Bigr)
  \;\le\;
  \E^A[Z]
  \;\le\;
  \Phi\Bigl(\log\frac{2}{\Pr[A]}\Bigr),
\]
where the first inequality holds because $Z\ge\Phi(\log(2/\delta))$
on $A$ by construction, and the second is $(\ast)$ applied to this
particular $A$. As $\Phi$ is strictly increasing (since $R,B>0$), this
forces $\Pr[A]\le\delta$, i.e.\ $Z\le\Phi(\log(2/\delta))$ with
probability at least $1-\delta$, from which the conclusion of the
Theorem is immediate. We therefore concentrate, without loss of
generality, on establishing $(\ast)$.

The proof of $(\ast)$ follows the chaining strategy
of~\cite[Appendix~A]{van_Handel_2010}, with a control on the realized
quadratic variation, which also implies the use of lower brackets in
the chaining mechanism.

\medskip\noindent
\textbf{Setup.}
Fix an event $A\in\mathcal F$ with $\Pr[A]>0$. Set
$\eta_j=2^{-j}\sqrt R$ and $N_j=N(\eta_j,\Theta,\hat\rho_n)$ for
$j\ge 0$. For each $j$, choose a bracketing family
$\mathcal B_j=\{(\ell_{j,p},u_{j,p})\}_{p=1}^{N_j}$ at radius
$\eta_j$. For each $\theta\in\Theta$ and level $j$, select an index
$\pi(j,\theta)$ such that, writing
$\Delta_{j,\theta,i}=u_{j,\pi(j,\theta),i}-\ell_{j,\pi(j,\theta),i}$
and $L_{j,\theta,i}=\ell_{j,\pi(j,\theta),i}$, we have
\[
  L_{j,\theta,i}\le X_i^\theta\le L_{j,\theta,i}+\Delta_{j,\theta,i},
  \qquad
  \hat\rho_n(\Delta_{j,\theta})\le\eta_j.
\]

\medskip\noindent
\textbf{Decomposition.}
Fix $J\ge 1$ and positive truncation levels $(a_j)_{0\le j\le J-1}$.
Define the stopping rule
$\tau_i^\theta=\min\{j\ge 0:\Delta_{j,\theta,i}>a_j\}\wedge J$,
which identifies the first level at which the bracket width exceeds
the truncation threshold. With the convention
$L_{-1,\theta,i}=L_{0,\theta,i}$, decompose
\[
  X_i^\theta
  \;=\;
  L_{0,\theta,i}
  +\sum_{j=0}^J b_{j,\theta,i}
  +\sum_{j=1}^J c_{j,\theta,i},
\]
where
\[
  b_{j,\theta,i}
  \;=\;
  \bigl(X_i^\theta-L_{j,\theta,i}\vee L_{j-1,\theta,i}\bigr)
    \mathbf 1_{\{\tau_i^\theta=j\}}
\]
is the residual at the stopping level, and
\[
  c_{j,\theta,i}
  \;=\;
  \bigl(L_{j,\theta,i}\vee L_{j-1,\theta,i}-L_{j-1,\theta,i}\bigr)
    \mathbf 1_{\{\tau_i^\theta=j\}}
  +
  \bigl(L_{j,\theta,i}-L_{j-1,\theta,i}\bigr)
    \mathbf 1_{\{\tau_i^\theta>j\}}
\]
is the chaining increment between consecutive bracket levels. The
decomposition telescopes: on $\{\tau_i^\theta = m\}$, the increments
$c_{k,\theta,i}$ for $k < m$ (where $\tau_i^\theta > k$) contribute
$L_{k,\theta,i}-L_{k-1,\theta,i}$, and at $k=m$ they contribute
$L_{m,\theta,i}\vee L_{m-1,\theta,i}-L_{m-1,\theta,i}$, so
$\sum_{k=1}^m c_{k,\theta,i}
   = L_{m,\theta,i}\vee L_{m-1,\theta,i} - L_{0,\theta,i}$;
adding $b_{m,\theta,i}$ gives $X_i^\theta - L_{0,\theta,i}$.
Passing to centered averages, $M_n^\theta=A_n^\theta
  +\sum_{j=0}^J B_{j,n}^\theta+\sum_{j=1}^J C_{j,n}^\theta$,
where each piece is the normalized martingale sum of the
corresponding array.

\medskip\noindent
\textbf{Control of $A_n^\theta$.}
On $\{\hat\rho_n(X^\theta)\le\sqrt R\}$, the base bracket satisfies
$n^{-1}\sum_i(L_{0,\theta,i})^2\le 4R$ (since
$(L_{0,\theta,i})^2\le 2(X_i^\theta)^2+2\Delta_{0,\theta,i}^2$ and
both empirical norms are at most $\sqrt R$). Since $A_n^\theta$
depends on $\theta$ only through $\pi(0,\theta)$, it takes at most
$N_0$ values. Corollary~\ref{cor:self-normalized-finite} with
realized variance $4R$ and envelope $B$ gives
\[
  \E^A\bigl[\sup_\theta
    \mathbf 1_{\{\hat\rho_n(X^\theta)\le\sqrt R\}}
    A_n^\theta\bigr]
  \;\le\;
  C\Bigl(\sqrt{\frac{R\,\ell_0}{n}}
    +\frac{B\,\ell_0}{n}\Bigr),
  \qquad
  \ell_0=\log\frac{2N_0}{\Pr[A]}.
\]

\medskip\noindent
\textbf{Control of the residuals $B_{j,n}^\theta$.}
Since $L_{j,\theta,i}\le X_i^\theta$ and
$L_{j-1,\theta,i}\le X_i^\theta$, we have
$L_{j,\theta,i}\vee L_{j-1,\theta,i}\le X_i^\theta$, so
$b_{j,\theta,i}\ge 0$, and
$b_{j,\theta,i}\le X_i^\theta-L_{j,\theta,i}\le\Delta_{j,\theta,i}$.
For $j<J$, since $\tau_i^\theta=j$ implies
$\Delta_{j,\theta,i}>a_j$, we have
$b_{j,\theta,i}\le\Delta_{j,\theta,i}^2/a_j$, so
$B_{j,n}^\theta\le\eta_j^2/a_j$. For $j=J$, Cauchy--Schwarz gives
$B_{J,n}^\theta\le\eta_J$. Hence
\[
  \sum_{j=0}^J
  \E^A\bigl[\sup_\theta
    \mathbf 1_{\{\hat\rho_n(X^\theta)\le\sqrt R\}}
    B_{j,n}^\theta\bigr]
  \;\le\;
  \sum_{j=0}^{J-1}\frac{\eta_j^2}{a_j}+\eta_J.
\]

\medskip\noindent
\textbf{Control of the chain increments $C_{j,n}^\theta$.}
Fix $j\ge 1$. The array $c_{j,\theta}$ depends on $\theta$ only
through $(\pi(0,\theta),\dots,\pi(j,\theta))$, so the supremum
reduces to a maximum over $\widetilde N_j=\prod_{p=0}^j N_p$ values.
We establish two pointwise bounds. First, both brackets trap
$X_i^\theta$, so
$|L_{j,\theta,i}-L_{j-1,\theta,i}|\le\Delta_{j-1,\theta,i}\vee
  \Delta_{j,\theta,i}$, and the positive part satisfies
$(L_{j,\theta,i}-L_{j-1,\theta,i})_+
   \le X_i^\theta-L_{j-1,\theta,i}\le\Delta_{j-1,\theta,i}$.
On $\{\tau_i^\theta=j\}$, we have $\tau_i^\theta>j-1$, so
$\Delta_{j-1,\theta,i}\le a_{j-1}$, giving
$c_{j,\theta,i}=(L_{j,\theta,i}-L_{j-1,\theta,i})_+\le a_{j-1}$.
On $\{\tau_i^\theta>j\}$, both
$\Delta_{j-1,\theta,i}\le a_{j-1}$ and
$\Delta_{j,\theta,i}\le a_j\le a_{j-1}$ hold (assuming the
$a_j$ are nonincreasing, which our choice below will satisfy), so
$|c_{j,\theta,i}|\le a_{j-1}$. For the empirical second moment,
on $\{\tau_i^\theta=j\}$ we have $c_{j,\theta,i}\le
\Delta_{j-1,\theta,i}$, and on $\{\tau_i^\theta>j\}$ we have
$|c_{j,\theta,i}|\le\Delta_{j-1,\theta,i}+\Delta_{j,\theta,i}$, so
$n^{-1}\sum_i(c_{j,\theta,i})^2\le 2(\eta_{j-1}^2+\eta_j^2)\le
  4\eta_{j-1}^2$. Corollary~\ref{cor:self-normalized-finite} with
$\ell_j=\log(2\widetilde N_j/\Pr[A])$ gives
\[
  \E^A\bigl[\sup_\theta
    \mathbf 1_{\{\hat\rho_n(X^\theta)\le\sqrt R\}}
    C_{j,n}^\theta\bigr]
  \;\le\;
  C\Bigl(\frac{\eta_{j-1}\sqrt{\ell_j}}{\sqrt n}
    +\frac{a_{j-1}\ell_j}{n}\Bigr).
\]

\medskip\noindent
\textbf{Choice of truncation levels and conclusion.}
Set $a_j=\eta_j\sqrt{n}/\sqrt{\ell_{j+1}}$ for $0\le j\le J-1$; this
sequence is nonincreasing since $\eta_j$ decreases geometrically
and $\ell_{j+1}$ increases. Then
$\eta_j^2/a_j=\eta_j\sqrt{\ell_{j+1}}/\sqrt{n}$ and
$a_{j-1}\ell_j/n=\eta_{j-1}\sqrt{\ell_j}/\sqrt{n}$. Combining all
three contributions,
\[
  \E^A\bigl[\sup_\theta
    \mathbf 1_{\{\hat\rho_n(X^\theta)\le\sqrt R\}}
    M_n^\theta\bigr]
  \;\le\;
  \eta_J
  +C\Bigl(
    \sqrt{\frac{R\,\ell_0}{n}}
    +\frac{B\,\ell_0}{n}
    +\frac{1}{\sqrt n}\sum_{j=0}^J\eta_j\sqrt{\ell_j}
  \Bigr).
\]
Letting $J\to\infty$ removes the $\eta_J$ term. Since
$\ell_j=\log(2/\Pr[A])+\sum_{p=0}^j\log N_p$ and
$\sqrt{x+y}\le\sqrt x+\sqrt y$, summing over $j$ with
$\sum_{j\ge 0}\eta_j=2\sqrt R$ and
$\sum_{j\ge p}\eta_j\le 2\eta_p$ yields
\[
  \frac{1}{\sqrt n}\sum_{j\ge 0}\eta_j\sqrt{\ell_j}
  \;\le\;
  C\Bigl(
    \sqrt{\frac{R}{n}\log\frac{2}{\Pr[A]}}
    +\frac{1}{\sqrt n}
      \sum_{p\ge 0}\eta_p\sqrt{\log N_p}
  \Bigr).
\]
Finally, since $N(\cdot,\Theta,\hat\rho_n)$ is nonincreasing, on
each interval $[\eta_{p+1},\eta_p]$ we have
$\sqrt{\log N(u)}\ge\sqrt{\log N_p}$, so
$\sum_{p\ge 0}\eta_p\sqrt{\log N_p}
  \le 2\int_0^{\sqrt R}\sqrt{\log N(u,\Theta,\hat\rho_n)}\,du$.
Substituting and using
$\ell_0\le\log(2/\Pr[A])+\log N(\sqrt R,\Theta,\hat\rho_n)$
gives $(\ast)$, and hence the Theorem.
\end{proof}
\newpage

\section{Complexity control and uniform confidence}
\label{app:complexity_control}

\subsection{Entropy transfer on the log scale
(Proposition~\ref{prop:entropy_equivalence})}
\label{app:transfer_inequality}

\begin{proposition}[Entropy transfer on the log scale]
Assume $Y_s\in[-1,0]$ almost surely and $\alpha\ge 1$. For every
$t\ge 1$, $\sigma>0$, and $\epsilon>0$,
\[
  N_{[\,]}\!\bigl(\epsilon,\,
    \mathcal L^\alpha_{1:t}(\sigma),\,\hat\rho_t\bigr)
  \;\le\;
  N\!\Bigl(\frac{c\,\epsilon}{1+\sigma},\,
    \log\Pi,\,\|\cdot\|_\infty\Bigr)
\]
for a universal constant $c>0$, where
$\mathcal L^\alpha_{1:t}(\sigma)
  =\bigl\{\pi\in\Pi:\widehat \rho_t( \ell_{1:t,\alpha}(\pi))\le\sigma\bigr\}$.
\end{proposition}

\begin{proof}
Fix a realization. Write
$w_s^\pi=\pi(A_s\mid X_s)/\pi_s(A_s\mid X_s)$ and
$\ell_s^\pi=(\min(w_s^\pi,\alpha)-1)\,Y_s$.
Let $\delta\in(0,1]$ and let $\{g^j\}_{j=1}^N$ be a $\delta$-cover
of $\log\Pi$ in $\|\cdot\|_\infty$. For each $\pi\in\Pi$, choose $j$
with $\|\log\pi-g^j\|_\infty\le\delta$, and set
$\widetilde w_s^j=\exp(g^j(X_s,A_s))/\pi_s(A_s\mid X_s)$. Then
$e^{-\delta}\widetilde w_s^j \le w_s^\pi \le e^{\delta}\widetilde w_s^j$
for every $s\le t$.

The map $h_s:w\mapsto(\min(w,\alpha)-1)\,Y_s$ is nonincreasing
(since $Y_s\le 0$) and $|Y_s|$-Lipschitz. Define brackets
$\lambda_s^j := h_s(e^{\delta}\widetilde w_s^j)$,
$\gamma_s^j := h_s(e^{-\delta}\widetilde w_s^j)$.
Monotonicity gives $\lambda_s^j\le\ell_s^\pi\le\gamma_s^j$.

We bound the bracket width directly in terms of clipped quantities,
avoiding any appeal to the unclipped empirical norm.  Write
$b_s:=e^{-\delta}\widetilde w_s^j\le w_s^\pi$, so the upper endpoint
is $e^{2\delta}b_s$.  Then
\[
  \gamma_s^j - \lambda_s^j
  = \bigl(\min(e^{2\delta}b_s,\alpha)
          - \min(b_s,\alpha)\bigr)\,|Y_s|.
\]
Since $x\mapsto\min(x,\alpha)$ is concave with
$\min(0,\alpha)=0$, it is sublinear:
$\min(e^{2\delta}b_s,\alpha)\le e^{2\delta}\min(b_s,\alpha)$.
Hence, using $b_s\le w_s^\pi$,
\[
  \gamma_s^j - \lambda_s^j
  \;\le\; (e^{2\delta}-1)\min(b_s,\alpha)\,|Y_s|
  \;\le\; (e^{2\delta}-1)\min(w_s^\pi,\alpha)\,|Y_s|
  \;\le\; C\delta\,\min(w_s^\pi,\alpha)\,|Y_s|,
\]
for $\delta\in(0,1]$ and some universal constant $C$.  Now observe
\[
  \min(w_s^\pi,\alpha)\,|Y_s|
  \;=\;
  \bigl|(\min(w_s^\pi,\alpha)-1)Y_s + Y_s\bigr|
  \;\le\; |\ell_s^\pi| + |Y_s|.
\]
Taking the empirical norm and applying the triangle inequality,
\[
  \hat\rho_t(\gamma^j-\lambda^j)
  \;\le\; C\delta\,\bigl(\hat\rho_t(\ell^\pi) + \hat\rho_t(|Y|)\bigr)
  \;\le\; C\delta(\sigma + 1),
\]
where the last step uses $\pi\in\mathcal L^\alpha_{1:t}(\sigma)$
(which gives $\hat\rho_t(\ell^\pi)\le\sigma$) and $|Y_s|\le 1$
(which gives $\hat\rho_t(|Y|)\le 1$).  Choosing
$\delta=c\,\epsilon/(1+\sigma)$ for a small universal constant $c$
completes the proof.
\end{proof}

\subsection{Bounding the complexity measure via entropy transfer}
\label{sec:bounding-complexity}

We apply Theorem~\ref{thm:selfnorm-chaining} to the clipped
importance-weighted loss class~\eqref{eq:clipped_loss}, where the
envelope satisfies $|\ell_{\alpha,t}(\pi)(z)|\le\alpha$.

\paragraph{Instantiation under the entropy assumption.}
In the small-radius regime ($\sigma\ll 1$), the rescaling factor
$c\sigma/(1+\sigma)\asymp\sigma$ in
Proposition~\ref{prop:entropy_equivalence} is negligible.
Substituting the entropy assumption
$\log N(\epsilon,\log\Pi,\|\cdot\|_\infty)\lesssim\epsilon^{-p}$
into the complexity measure of Theorem~\ref{thm:selfnorm-chaining}
gives the upper bound
\begin{equation}\label{eq:H-tilde}
  \mathcal{H}_t(\sigma,\alpha)
  \;:=\;
  \frac{\alpha}{t}\,\sigma^{-p}
  \;+\;
  \frac{\sigma^{1-p/2}}{\sqrt{t}}.
\end{equation}
Solving the fixed-point equation
$\mathcal{H}_t(r,\alpha)=r^2$ from~\eqref{eq:critical_radius_abstract}
yields the critical radius
$r_{\alpha,t}\asymp(\alpha_t/t)^{1/(2+p)}$.

\paragraph{The minimal point.}
The two terms in~\eqref{eq:H-tilde} balance at the scale
\begin{equation}\label{eq:inflection}
  r_t^*
  \;\asymp\;
  \Bigl(\frac{\alpha_t}{\sqrt{t}}\Bigr)^{\!\frac{2}{2+p}}.
\end{equation}
Below $r_t^*$, the entropy term $\alpha\sigma^{-p}/t$ dominates;
above it, the Dudley-integral term $\sigma^{1-p/2}/\sqrt{t}$
dominates. For $\sigma\gg 1$, the rescaling saturates and
$\mathcal{H}_t(\sigma)\lesssim\sigma/\sqrt{t}$.

\paragraph{Two-scale control.}
The following lemma collects the behaviour of $\mathcal{H}_t$ across
the full range of localization radii.

\begin{lemma}[Two-scale control]
\label{lem:H_two_scale}
Under the entropy assumption~\ref{asm:entropy_logPi_scale} with
$p\in(0,2)$, suppose $r_t^*\le c_0$ for a sufficiently small
universal constant. Then:
\begin{enumerate}[label=(\roman*),nosep]
  \item For $\sigma\in(0,c_0]$:
    $\mathcal{H}_t(\sigma)
      \lesssim \alpha_t\sigma^{-p}/t + \sigma^{1-p/2}/\sqrt{t}$.
  \item For $\sigma\in[r_{\alpha,t},r_t^*]$:
    $\mathcal{H}_t(\sigma)\lesssim(r_{\alpha,t})^2$.
  \item For $\sigma\in[r_t^*,c_0]$:
    $\mathcal{H}_t(\sigma)\lesssim\sigma^{1-p/2}/\sqrt{t}$.
  \item For $\sigma\ge c_0$:
    $\mathcal{H}_t(\sigma)\lesssim\sigma/\sqrt{t}$.
\end{enumerate}
\end{lemma}

\begin{proof}
Part~(i) is immediate from~\eqref{eq:H-tilde}. For~(ii), on
$[r_{\alpha,t},r_t^*]$ the entropy term satisfies
$\alpha_t\sigma^{-p}/t \le \alpha_t(r_{\alpha,t})^{-p}/t
  = (r_{\alpha,t})^2$,
and the Dudley term satisfies
$(r_t^*)^{1-p/2}/\sqrt{t} \asymp \alpha_t(r_t^*)^{-p}/t
  \lesssim (r_{\alpha,t})^2$.
For~(iii), when $\sigma\ge r_t^*$ the entropy term is dominated by
the Dudley term. Part~(iv) follows from the saturation of the
rescaling factor at $\sigma\gg 1$.
\end{proof}

\subsection{Self-normalized bound via peeling
(Theorem~\ref{thm:peeling})}
\label{sec:peeling}

We combine Theorem~\ref{thm:selfnorm-chaining} with
Lemma~\ref{lem:H_two_scale} via a dyadic peeling argument over the
shells of $\hat\rho_t(\ell_{\alpha,1:t}(\pi))$.

\begin{theorem}[Self-normalized bound via peeling]
\label{thm:peeling-appendix}
Under the entropy condition~\eqref{asm:entropy_logPi_scale} with
$p\in(0,2)$, there exists $C>0$ such that with probability at least
$1-\eta$, every $\pi\in\Pi$ satisfies
\[
  M_t^\pi
  \;\le\;
  C\Bigg(
    (r_{\alpha,t})^2
    + \frac{\hat\rho_t(\ell_{\alpha,1:t}(\pi))^{1-p/2}}{\sqrt{t}}
    + \hat\rho_t(\ell_{\alpha,1:t}(\pi))
      \sqrt{\frac{L_t(\eta)}{t}}
    + \frac{\alpha_t}{t}\,L_t(\eta)
  \Bigg),
\]
where $L_t(\eta) := \log(\alpha_t/r_t^*)+\log(1/\eta)$.
\end{theorem}

\begin{proof}
Recall the localized deviation envelope from
Theorem~\ref{thm:selfnorm-chaining}: for any $\sigma>0$ and
$\eta\in(0,1)$,
\begin{equation}\label{eq:eps-sigma}
  \varepsilon_t(\sigma,\eta)
  \;:=\;
  C\Bigg(
    \mathcal{H}_t(\sigma)
    + \sigma\sqrt{\frac{\log(2/\eta)}{t}}
    + \frac{\alpha_t}{t}\log\frac{2}{\eta}
  \Bigg).
\end{equation}

\emph{Shell~1} ($\hat\rho_t\le r_{\alpha,t}$). Apply~\eqref{eq:eps-sigma}
at $r_{\alpha,t}$ with confidence $\eta/3$. The critical-radius property
gives $\varepsilon_t(r_{\alpha,t},\eta/3)\lesssim(r_{\alpha,t})^2
  + (\alpha_t/t)L_t(\eta)$.

\emph{Shell~2} ($r_{\alpha,t}<\hat\rho_t\le r_t^*$).
Apply~\eqref{eq:eps-sigma} at $r_t^*$ with confidence $\eta/3$.
Lemma~\ref{lem:H_two_scale}(ii) gives
$\mathcal{H}_t(\sigma)\lesssim(r_{\alpha,t})^2$ throughout this range.

\emph{Shell~3} ($r_t^*<\hat\rho_t\le\alpha_t$). Peel dyadically:
$S_k=\{2^{k-1}r_t^*<\hat\rho_t\le 2^k r_t^*\}$ for
$k=1,\dots,K$ with $K=\lceil\log_2(\alpha_t/r_t^*)\rceil$ and
confidence $\eta_k=\eta/(3(k+1)^2)$ per shell. On $S_k$ with
$\sigma_k=2^k r_t^*$:
\begin{itemize}[nosep]
  \item If $\sigma_k\le c_0$:
    Lemma~\ref{lem:H_two_scale}(iii) gives
    $\mathcal{H}_t(\sigma_k)\lesssim\sigma_k^{1-p/2}/\sqrt{t}$,
    contributing $\hat\rho_t^{1-p/2}/\sqrt{t}$.
  \item If $\sigma_k>c_0$:
    Lemma~\ref{lem:H_two_scale}(iv) gives
    $\mathcal{H}_t(\sigma_k)\lesssim\sigma_k/\sqrt{t}$, absorbed by
    the deviation term.
\end{itemize}
The union bound over $K$ shells costs $\log(\alpha_t/r_t^*)$,
which is absorbed into $L_t(\eta)$. Collecting the three
contributions completes the proof.
\end{proof}

\subsection{Verification of Condition~\ref{asm:emp_to_pred_var}}
\label{app:emp_to_pred}

The key step is to transfer from the empirical clipped norm
$\hat\rho_t=\hat\rho_t(\ell_{\alpha,1:t}(\pi^\star))$ to the
predictable variance $\bar\sigma_t=\bar\sigma_t(\pi^\star)$. Since
$\pi^\star$ is a single policy, no uniformity is needed.

\begin{proof}
Set
\[
  X_s:=\bigl(\min(w_s^{\pi^\star},\alpha_t)-1\bigr)^2,
  \qquad
  D_s:=X_s-\E[X_s\mid\mathcal F_{s-1}],
\]
so that $(D_s)_{s\le t}$ is a martingale-difference sequence with
\[
  \hat\sigma_t^2 - \bar\sigma_{t,{\rm clip}}^2
  =
  \frac1t\sum_{s=1}^t D_s,
  \qquad
  \bar\sigma_{t,{\rm clip}}^2
  :=
  \frac1t\sum_{s=1}^t
    \E\bigl[(\min(w_s^{\pi^\star},\alpha_t)-1)^2\mid\mathcal F_{s-1}\bigr].
\]
Note that this definition targets the clipped predictable variance
directly, bypassing the factor $Y_s^2$: in particular,
$\hat\sigma_t^2=t^{-1}\sum_s X_s$ is precisely the empirical clipped
variance appearing in Condition~\ref{asm:emp_to_pred_var}.

By clipping, $\min(w_s^{\pi^\star},\alpha_t)\in[0,\alpha_t]$, hence
$(\min(w_s^{\pi^\star},\alpha_t)-1)^2\le\alpha_t^2$ (using
$\alpha_t\ge 1$), so
\[
  0\le X_s\le \alpha_t^2,
  \qquad
  |D_s|\le \alpha_t^2.
\]
Moreover, clipping contracts toward $1$, so
$(\min(w,\alpha_t)-1)^2\le(w-1)^2$ pointwise; therefore
\[
  \bar\sigma_{t,{\rm clip}}^2
  \le
  \frac1t\sum_{s=1}^t
    \E\bigl[(w_s^{\pi^\star}-1)^2\mid\mathcal F_{s-1}\bigr]
  =
  \bar\sigma_t^2.
\]
For the conditional second moment of $D_s$,
\[
  \E[D_s^2\mid\mathcal F_{s-1}]
  \le
  \E[X_s^2\mid\mathcal F_{s-1}]
  \le
  \alpha_t^2 \E[X_s\mid\mathcal F_{s-1}],
\]
since $X_s^2\le\alpha_t^2 X_s$.  Setting
$V_t:=\sum_{s=1}^t\E[D_s^2\mid\mathcal F_{s-1}]$,
\[
  V_t
  \le
  \alpha_t^2\sum_{s=1}^t\E[X_s\mid\mathcal F_{s-1}]
  =
  t\alpha_t^2\bar\sigma_{t,{\rm clip}}^2
  \le
  t\alpha_t^2\bar\sigma_t^2.
\]
Applying Freedman's inequality to $\sum_{s=1}^t D_s$ with increment
bound $\alpha_t^2$ yields, with probability at least $1-\delta_t$,
\[
  \sum_{s=1}^t D_s
  \lesssim
  \sqrt{V_t\log(1/\delta_t)}
  +\alpha_t^2\log(1/\delta_t).
\]
Using the bound on $V_t$ and dividing by $t$,
\[
  \hat\sigma_t^2
  \lesssim
  \bar\sigma_t^2
  +\alpha_t\bar\sigma_t\sqrt{\frac{\log(1/\delta_t)}{t}}
  +\alpha_t^2\,\frac{\log(1/\delta_t)}{t}.
\]
Taking square roots,
\begin{equation}\label{eq:key-compare}
  \hat\sigma_t
  \le
  c_1(\bar\sigma_t+\psi_t),
  \qquad
  \psi_t:=c_0\alpha_t\sqrt{\frac{\log(1/\delta_t)}{t}}.
\end{equation}
Condition~\ref{asm:HEB} controls the predictable scale: because
$\bar\sigma_t^2$ is bounded by a power of the excess risk and the
latter is uniformly bounded, $\bar\sigma_t=O(1)$.  Thus the
predictable scale of the clipped process remains uniformly
controlled.  Clipping is essential: it yields the bounded-increment
condition and the bound on $V_t$ needed for Freedman's inequality.

\medskip\noindent
\textbf{From \eqref{eq:key-compare} to a penalty bound.}
We now work on the intersection of the concentration event
\eqref{eq:key-compare} and the uniform confidence event, and bound
$\Phi(\hat\sigma_t)$ directly, without case-splitting on the
ordering of $\psi_t$ and $r_{\alpha,t}$.

The penalty $\Phi$ takes the form
$\Phi(\sigma)\asymp r_{\alpha,t}^2+\sigma^{1-p/2}/\sqrt t
  +\sigma\sqrt{L_t/t}+\alpha_t L_t/t$,
which (away from the floor $r_{\alpha,t}^2$) is a sum of terms of
the form $\sigma^q$ with $q\in\{1-p/2,\,1\}$, $q\in(0,1]$ for
$p\in[0,2)$.  On $[0,\infty)$, the map $x\mapsto x^q$ is
subadditive for $q\in(0,1]$:
$(A+B)^q\le A^q+B^q$.  Consequently,
\[
  \Phi\bigl(c_1(\bar\sigma_t+\psi_t)\bigr)
  \;\le\;
  C\bigl(\Phi(\bar\sigma_t)+\Phi(\psi_t)\bigr)
\]
for a universal constant $C$ depending on $c_1$ and the exponents
in $\Phi$.  Combined with \eqref{eq:key-compare} and the
monotonicity of $\Phi$ on $[r_{\alpha,t},\infty)$,
\begin{equation}\label{eq:phi-split}
  \Phi(\hat\sigma_t)
  \;\le\;
  \Phi\bigl(c_1(\bar\sigma_t+\psi_t)\bigr)
  \;\le\;
  C\bigl(\Phi(\bar\sigma_t)+\Phi(\psi_t)\bigr).
\end{equation}

\medskip\noindent
\textbf{Absorption of $\Phi(\psi_t)$ into $r_{\alpha,t}^2$.}
With $\psi_t\asymp\alpha_t\sqrt{\log(1/\delta_t)/t}$,
\[
  \Phi(\psi_t)
  \;\asymp\;
  r_{\alpha,t}^2
  +\frac{\psi_t^{1-p/2}}{\sqrt t}
  +\psi_t\sqrt{\frac{L_t}{t}}
  +\alpha_t\frac{L_t}{t}.
\]
The last two terms are absorbed into $r_{\alpha,t}^2$ by the same
reasoning used below for $\bar\sigma_t$.  For the dominant term,
\[
  \frac{\psi_t^{1-p/2}}{\sqrt t}
  \;\asymp\;
  \alpha_t^{1-p/2}\,t^{(p-4)/4}\,(\log(1/\delta_t))^{(1-p/2)/2},
\]
up to polylogarithmic factors.  A direct comparison of exponents
with $r_{\alpha,t}^2\asymp(\alpha_t/t)^{4/(4+p)}$ shows that, for
every $p\in[0,2)$ and every polynomial schedule
$\alpha_t\asymp t^\gamma$ with
$\gamma\ge p/(2(2+p))$ in the range admitted by
Theorem~\ref{thm:regret}, one has
$\Phi(\psi_t)\lesssim r_{\alpha,t}^2$.

Substituting into \eqref{eq:phi-split},
\[
  \Phi(\hat\sigma_t)
  \;\lesssim\;
  \Phi(\bar\sigma_t)+r_{\alpha,t}^2.
\]
Expanding $\Phi(\bar\sigma_t)$, using $\bar\sigma_t=O(1)$, and
absorbing the lower-order terms
$\bar\sigma_t\sqrt{L_t/t}+\alpha_t L_t/t$ into
$r_{\alpha,t}^2+\bar\sigma_t^{1-p/2}/\sqrt t$ yields
\[
  \Phi(\hat\sigma_t)
  \;\lesssim\;
  r_{\alpha,t}^2
  +\frac{\bar\sigma_t^{\,1-p/2}}{\sqrt t}.
\]
This ends the proof.
\end{proof}

\section{Regret analysis}
\label{app:regret_analysis}
In this section, we prove the master theorem, using the upper bound of the excess risk derived from conditions ~\ref{asm:surrogate_risk}--\ref{asm:heb}.

\begin{proof}[Proof of Theorem~\ref{thm:regret}]
Assume that, on the high-probability event from Condition~\ref{asm:surrogate_risk}, the one-step excess risk satisfies \begin{equation}\label{eq:uniform-master-recursion} R(\pi_{t+1})-R(\pi^\star) \;\lesssim\; \frac{\bar\sigma_t^{\,1-b}}{\sqrt t} \;+\; \Bigl(\frac{\alpha_t}{t}\Bigr)^{\!\frac{1}{1+b}} \;+\; \frac{\bar\sigma_t^2}{\alpha_t}, \end{equation} where \[ \bar\sigma_t^2:=\frac1t\sum_{s=1}^t \sigma_s^2. \] (As usual, the small-variance branch is absorbed into the critical-radius term $(\alpha_t/t)^{1/(1+b)}$.)
Write
\[
  e_t:=R(\pi_t)-R(\pi^\star),
  \qquad
  F_t:=\sum_{s=1}^t e_s
\]
By Condition~\ref{asm:heb}, $\sigma_s^2\lesssim e_s^\beta$. Since
$u\mapsto u^\beta$ is concave on $[0,\infty)$ for $\beta\in(0,1]$,
Jensen's inequality gives
\begin{equation}\label{eq:jensen-sigma-corrected}
  \bar\sigma_t^2
  \;\lesssim\;
  \frac1t\sum_{s=1}^t e_s^\beta
  \;\le\;
  \Bigl(\frac{F_t}{t}\Bigr)^\beta.
\end{equation}

Let
\[
  E:=\bigcap_{t\ge1} E_t,
\]
where $E_t$ is the high-probability event from
Condition~\ref{asm:surrogate_risk} at round $t$. Since
$\sum_{t\ge1}\delta_t\le \delta$, a union bound yields
$\Pr(E)\ge 1-\delta$. We work on $E$ throughout.

We prove the claimed rates on $E$.

\paragraph{Case 1: $\beta\in(0,1)$.}
Set
\[
  a:=\frac{1}{2+b-\beta},
  \qquad
  c:=\frac{1-\beta}{2+b-\beta},
  \qquad
  \alpha_t\asymp t^c.
\]
We show that
\[
  F_t=\widetilde{\mathcal O}\!\bigl(t^{1-a}\bigr),
\]
which immediately implies
\[
  e_t=\widetilde{\mathcal O}(t^{-a})
  \qquad\text{and}\qquad
  \mathrm{Regret}_T=\widetilde{\mathcal O}(T^{1-a}).
\]

To keep notation light, let $L_t$ denote a generic nondecreasing
polylogarithmic factor in $t$ and $1/\delta$; its value may change from
line to line.

Assume inductively that
\[
  F_t \lesssim t^{1-a}L_t.
\]
Then by~\eqref{eq:jensen-sigma-corrected},
\[
  \bar\sigma_t^2
  \lesssim
  t^{-a\beta} L_t^\beta,
  \qquad
  \bar\sigma_t
  \lesssim
  t^{-a\beta/2} L_t^{\beta/2}.
\]
Substituting into~\eqref{eq:uniform-master-recursion} gives
\[
  e_{t+1}
  \lesssim
  t^{-E_1}L_t^{\beta(1-b)/2}
  +
  t^{-E_2}
  +
  t^{-E_3}L_t^\beta,
\]
where
\[
  E_1:=\frac12+\frac{a\beta(1-b)}2,
  \qquad
  E_2:=\frac{1-c}{1+b},
  \qquad
  E_3:=a\beta+c.
\]
With the above choice of $a,c$,
\[
  E_2
  =
  \frac{1-c}{1+b}
  =
  \frac{1}{2+b-\beta}
  =
  a,
\]
and
\[
  E_3
  =
  a\beta+c
  =
  \frac{\beta}{2+b-\beta}
  +
  \frac{1-\beta}{2+b-\beta}
  =
  a.
\]
Moreover,
\[
  E_1-a
  =
  \frac12+\frac{a\beta(1-b)}2-a
  =
  \frac{b(1-\beta)}{2(2+b-\beta)}
  \ge 0.
\]
Hence all three terms are bounded by
$\widetilde{\mathcal O}(t^{-a})$, and therefore
\[
  e_{t+1}
  =
  \widetilde{\mathcal O}(t^{-a}).
\]
Summing from $1$ to $T$ yields
\[
  F_T
  =
  \sum_{t=1}^T e_t
  =
  \widetilde{\mathcal O}\!\Bigl(\sum_{t=1}^T t^{-a}\Bigr)
  =
  \widetilde{\mathcal O}\!\bigl(T^{1-a}\bigr),
\]
since $a<1$. This closes the induction and proves
\[
  e_t=\widetilde{\mathcal O}\!\Bigl(t^{-\frac{1}{2+b-\beta}}\Bigr),
  \qquad
  \mathrm{Regret}_T
  =
  \widetilde{\mathcal O}\!\Bigl(
    T^{\frac{1+b-\beta}{2+b-\beta}}
  \Bigr).
\]

\paragraph{Case 2: $\beta=1$ and $b\in(0,1]$.}
Set
\[
  a:=\frac{1}{1+b},
  \qquad
  \alpha_t:=1+\log(et).
\]
Again let $L_t$ denote a generic nondecreasing polylogarithmic factor.
Assume inductively that
\[
  F_t\lesssim t^{1-a}L_t.
\]
Then by~\eqref{eq:jensen-sigma-corrected},
\[
  \bar\sigma_t^2\lesssim t^{-a}L_t,
  \qquad
  \bar\sigma_t\lesssim t^{-a/2}L_t^{1/2}.
\]
Substituting into~\eqref{eq:uniform-master-recursion},
\begin{align*}
  e_{t+1}
  &\lesssim
  \frac{\bar\sigma_t^{\,1-b}}{\sqrt t}
  +
  \Bigl(\frac{\alpha_t}{t}\Bigr)^{\!\frac{1}{1+b}}
  +
  \frac{\bar\sigma_t^2}{\alpha_t} \\
  &\lesssim
  t^{-\frac12-\frac{a(1-b)}2} L_t^{(1-b)/2}
  +
  t^{-\frac{1}{1+b}}\alpha_t^{1/(1+b)}
  +
  t^{-a}\frac{L_t}{\alpha_t}.
\end{align*}
Since $a=1/(1+b)$,
\[
  \frac12+\frac{a(1-b)}2
  =
  \frac{1}{1+b}
  =
  a,
\]
so each term is $\widetilde{\mathcal O}(t^{-a})$. Hence
\[
  e_{t+1}
  =
  \widetilde{\mathcal O}(t^{-a}).
\]
Summing yields
\[
  F_T
  =
  \widetilde{\mathcal O}\!\bigl(T^{1-a}\bigr)
  =
  \widetilde{\mathcal O}\!\bigl(T^{\frac{b}{1+b}}\bigr),
\]
which proves
\[
  e_t=\widetilde{\mathcal O}\!\bigl(t^{-1/(1+b)}\bigr),
  \qquad
  \mathrm{Regret}_T=\widetilde{\mathcal O}\!\bigl(T^{b/(1+b)}\bigr).
\]

\paragraph{Case 3: $\beta=1$ and $b=0$.}
Set $\alpha_t:=1+\log(et)$. Then
\eqref{eq:uniform-master-recursion} becomes
\[
  e_{t+1}
  \lesssim
  \frac{\bar\sigma_t}{\sqrt t}
  +
  \frac{\alpha_t}{t}
  +
  \frac{\bar\sigma_t^2}{\alpha_t}.
\]
By~\eqref{eq:jensen-sigma-corrected},
\[
  \bar\sigma_t^2\lesssim \frac{F_t}{t}.
\]
Hence
\begin{equation}\label{eq:parametric-recursion}
  e_{t+1}
  \lesssim
  \frac{\sqrt{F_t}}{t}
  +
  \frac{\alpha_t}{t}
  +
  \frac{F_t}{t\,\alpha_t}.
\end{equation}
We claim that $F_t=\mathcal O(\log^2 t)$. Indeed, if
$F_t\le A\log^2(et)$ for all $t\le T$, then
\eqref{eq:parametric-recursion} gives
\[
  e_{t+1}
  \lesssim
  \frac{\log(et)}{t}.
\]
Summing over $t\le T$ yields
\[
  F_T
  =
  \sum_{t=1}^T e_t
  \lesssim
  \sum_{t=1}^T \frac{\log(et)}{t}
  =
  \mathcal O(\log^2 T),
\]
so the bootstrap closes for $A$ large enough.
Therefore
\[
  \mathrm{Regret}_T=\mathcal O(\log^2 T).
\]
The one-step bound \eqref{eq:parametric-recursion} then implies
\[
  e_t=\mathcal O\!\Bigl(\frac{\log(et)}{t}\Bigr),
\]
and in particular
\[
  e_t=\mathcal O\!\Bigl(\frac{\log^2 t}{t}\Bigr),
\]
which matches the stated theorem.

\paragraph{Case 4: $\beta \in (0,1)$, $b = 0$.}
Set $a := 1/(2-\beta)$, $\gamma := (1-\beta)/(2-\beta)$, and
$\alpha_t := 2 \vee t^{\gamma}$. Then
\eqref{eq:uniform-master-recursion} with $b=0$ becomes
\[
  e_{t+1}
  \;\lesssim\;
  \frac{\bar\sigma_t}{\sqrt{t}}
  + \frac{\alpha_t}{t}
  + \frac{\bar\sigma_t^2}{\alpha_t}.
\]
Assume inductively $F_t \le C\, t^{1-a} L_t$. By~\eqref{eq:jensen-sigma-corrected},
$\bar\sigma_t \lesssim t^{-a\beta/2} L_t^{\beta/2}$. The three terms
decay as $t^{-E_1}$, $t^{-E_2}$, $t^{-E_3}$ with
\[
  E_1 = \tfrac{1}{2} + \tfrac{a\beta}{2},
  \qquad
  E_2 = 1 - \gamma = a,
  \qquad
  E_3 = a\beta + \gamma = a.
\]
Since $E_1 - a = (1 - a(2-\beta))/2 = 0$, all three exponents equal $a$,
so $e_{t+1} = \widetilde{\mathcal{O}}(t^{-a})$. Summation gives
\[
  \mathrm{Regret}_T
  = \widetilde{\mathcal{O}}\!\left(T^{\frac{1-\beta}{2-\beta}}\right).
\]
\end{proof}

\newpage

\subsection{Clipping bias (Lemma~\ref{lem:clipping_bias})}
\label{app:clipping_bias}

\begin{proof}[Proof of Lemma~\ref{lem:clipping_bias}]
For $s\le t$, write
\[
\rho_s^\pi:=\frac{\pi(A_s\mid X_s)}{\pi_s(A_s\mid X_s)}.
\]
Then
\[
R_t(\pi)
=\frac1t\sum_{s=1}^t
\E\!\left[(\rho_s^\pi-1)Y_s\mid\mathcal F_{s-1}\right],
\qquad
R_{t,\alpha}(\pi)
=\frac1t\sum_{s=1}^t
\E\!\left[(\min(\rho_s^\pi,\alpha)-1)Y_s\mid\mathcal F_{s-1}\right].
\]

Since $\min(\rho_s^\pi,\alpha)\le \rho_s^\pi$ and $Y_s\le 0$,
\[
R_t(\pi)\le R_{t,\alpha}(\pi).
\]

For the upper bound, using
\[
\min(\rho_s^\pi,\alpha)=\rho_s^\pi-(\rho_s^\pi-\alpha)_+,
\]
we get
\[
R_{t,\alpha}(\pi)-R_t(\pi)
=
-\frac1t\sum_{s=1}^t
\E\!\left[(\rho_s^\pi-\alpha)_+Y_s\mid\mathcal F_{s-1}\right].
\]
Because $Y_s\in[-1,0]$,
\[
-(\rho_s^\pi-\alpha)_+Y_s\le (\rho_s^\pi-\alpha)_+,
\]
hence
\[
R_{t,\alpha}(\pi)-R_t(\pi)
\le
\frac1t\sum_{s=1}^t
\E\!\left[(\rho_s^\pi-\alpha)_+\mid\mathcal F_{s-1}\right].
\]

Now, for any nonnegative $Z$,
\[
\E[Z\mid\mathcal F]=\int_0^\infty \Pr(Z\ge u\mid\mathcal F)\,du.
\]
Applying this to $Z=(\rho_s^\pi-\alpha)_+$,
\[
\E\!\left[(\rho_s^\pi-\alpha)_+\mid\mathcal F_{s-1}\right]
=
\int_\alpha^\infty
\Pr(\rho_s^\pi\ge v\mid\mathcal F_{s-1})\,dv.
\]
For $v\ge \alpha>1$,
\[
\{\rho_s^\pi\ge v\}\subseteq \{(\rho_s^\pi-1)^2\ge (v-1)^2\},
\]
so conditional Markov's inequality gives
\[
\Pr(\rho_s^\pi\ge v\mid\mathcal F_{s-1})
\le
\frac{\E[(\rho_s^\pi-1)^2\mid\mathcal F_{s-1}]}{(v-1)^2}
=
\frac{\sigma_s^2(\pi)}{(v-1)^2}.
\]
Therefore,
\[
\E\!\left[(\rho_s^\pi-\alpha)_+\mid\mathcal F_{s-1}\right]
\le
\sigma_s^2(\pi)\int_\alpha^\infty \frac{dv}{(v-1)^2}
=
\frac{\sigma_s^2(\pi)}{\alpha-1}.
\]
Averaging over $s=1,\dots,t$ yields
\[
R_{t,\alpha}(\pi)
\le
R_t(\pi)+\frac{1}{t(\alpha-1)}\sum_{s=1}^t \sigma_s^2(\pi).
\]
Combining this with the lower bound proves \eqref{eq:clipping_bias_bound}.
\end{proof}

\section{Additional numerical experiments and details}
\label{app:additional_numerical}

In this section, we provide additional details about the experimental setup. All computations were performed on CPU; no GPUs, clusters, or cloud resources were used.

\subsection{ERM vs.\ SVP}
\label{subsec:erm_svp}

We evaluate SVP \citep{maurer2009empirical} under temporally dependent data using a Markov-chain design with $K = 500$ arms. For each coordinate $k$, generate $a_k \sim \mathrm{Unif}[B, 1-B]$ with $B = 0.25$, $b_k \sim \mathrm{Unif}[0, B]$, and autocorrelation $\chi_k \sim \mathrm{Unif}[0, \chi_{\max}]$ where $\chi_{\max} = 0.9$.

Data follows $X_{t,k} = a_k + b_k \cdot S_{t,k}$ where $(S_{t,k})$ is a Markov chain with $S_{0,k} \sim \mathrm{Unif}\{\pm 1\}$ and persistence probability $p_k = \frac{1 + \chi_k}{2}$, yielding $\mathrm{Cov}(X_{t,k}, X_{t+1,k}) = \chi_k b_k^2$.

We compare ERM ($\arg\min_k \widehat{\mu}_k$) against dependence-aware SVP:
\[
\arg\min_k \left[ \widehat{\mu}_k + \lambda \sqrt{\frac{\widehat{v}_k}{n_{\mathrm{eff},k}}} \right], \quad n_{\mathrm{eff},k} = n \cdot \frac{1 - \widehat{\chi}_k}{1 + \widehat{\chi}_k}
\]
with $\lambda = 2.5$ and $\widehat{\chi}_k$ the sample autocorrelation at lag 1. Performance is measured by excess risk $a_{\hat{k}} - \min_k a_k$ averaged over 1000 trials for $n \in \{50, 70, \ldots, 500\}$.

\subsection{Sensitivity to overlap in offline contextual bandits}
\label{app:offline_overlap}

To study how offline learning degrades as overlap becomes weaker, we consider a synthetic contextual bandit environment with binary rewards and vary only the behavior policy. More precisely, we use a context-dependent logger of the form
\[
\pi_b(a \mid x)
=
(1-\varepsilon)\,\tilde{\pi}_b(a \mid x)
+
\frac{\varepsilon}{K},
\]
where $\tilde{\pi}_b(\cdot \mid x)$ is a sharp contextual softmax policy and $K$ is the number of actions. The parameter $\varepsilon$ controls the amount of uniform exploration. Therefore the logging policy satisfies the uniform lower bound
\[
\pi_b(a \mid x) \ge \frac{\varepsilon}{K}
\qquad \text{for all } x,a,
\]
so decreasing $\varepsilon$ directly weakens overlap.

We report results as a function of the guaranteed overlap floor $\varepsilon / K$. In addition to exact policy value, we also show simple overlap diagnostics computed from the full behavior policy, namely the mean minimum action probability, the mean maximum action probability, and the mean entropy of $\pi_b(\cdot \mid x)$. These diagnostics confirm that as the overlap floor decreases, the logger becomes increasingly concentrated on a small subset of actions.

\begin{figure}[t]
    \centering
    \includegraphics[width=0.92\linewidth]{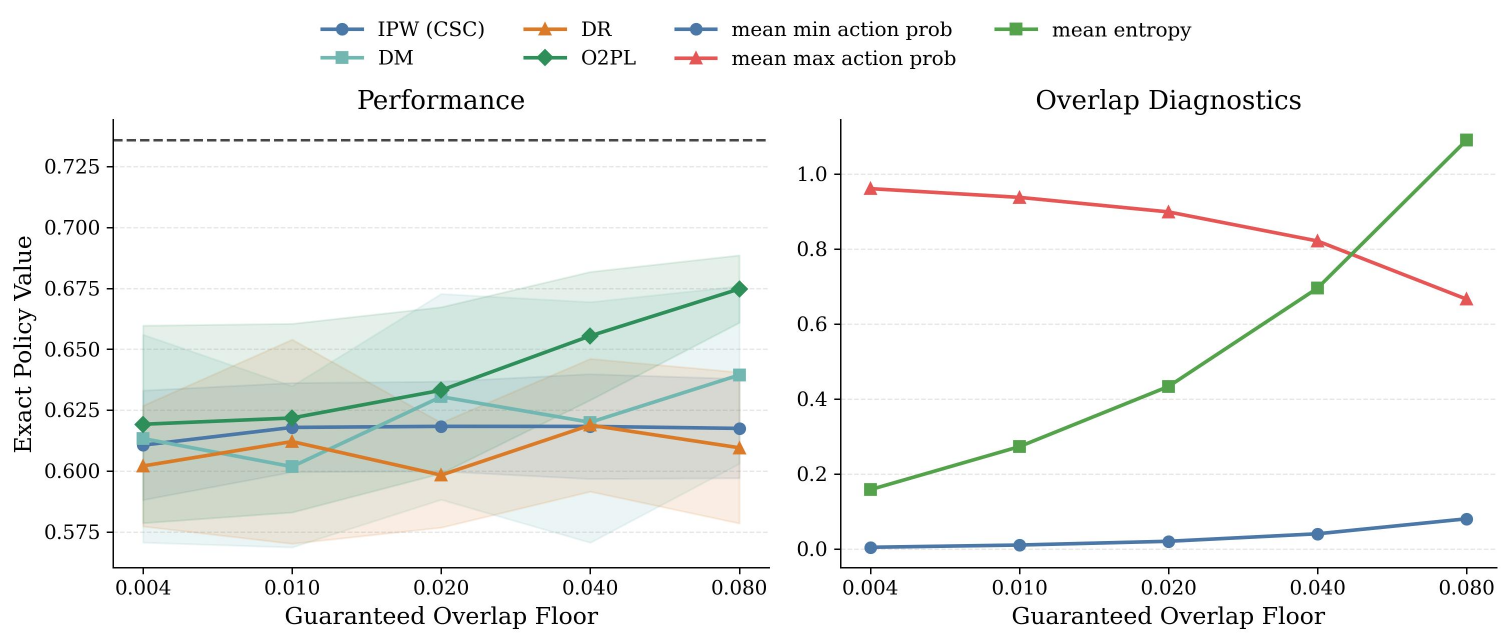}
    \caption{Sensitivity to overlap in offline contextual bandits. The left panel reports exact policy value as the guaranteed overlap floor $\varepsilon / K$ decreases, where smaller values correspond to weaker overlap. The right panel reports overlap diagnostics for the same logging policies: the mean minimum action probability, the mean maximum action probability, and the mean entropy of the logger. As overlap weakens, the logger becomes more concentrated and all methods deteriorate, but O2PL remains the strongest method across the sweep.}
    \label{fig:offline_overlap_strength}
\end{figure}

Figure~\ref{fig:offline_overlap_strength} shows that all methods deteriorate as overlap weakens, but O2PL remains the strongest method across the full sweep. The gap is clearest at moderate overlap levels, where clipping still stabilizes the objective without removing too much information from the logged sample. As the overlap floor becomes extremely small, the problem becomes difficult for all methods, but O2PL still degrades more gracefully than the standard IPW, direct-method, and doubly robust baselines.

\subsection{Comparison to agnostic methods}
\label{app:open_ml}

We evaluate O2PL on a benchmark of 100 OpenML multiclass datasets converted into contextual bandit problems, using 10 random seeds and a horizon capped at $T=5000$ on each dataset. Rewards are shifted by $+1$ for all learners. For O2PL, we use the flat replay-Adam variant with learning rate $0.005$, replay window and batch size $200$, and warm start $50$. Concretely, O2PL stores the most recent 200 interactions, and after the warm start it performs 5 Adam ascent steps every 5 interactions on the clipped-ratio pessimistic objective. The competing methods are off-the-shelf Vowpal Wabbit contextual-bandit learners using the feature map \texttt{a,ax}: Greedy, Cover-nu with ensemble size 3, Bag-greedy with ensemble size 3, and RegCB-opt in optimistic mode.

The datasets used in our experiments can be accessed at \url{https://www.openml.org/d/<id>}, where \texttt{<id>} is one of the following OpenML data IDs:
\begin{quote}\small
3, 6, 10, 11, 12, 14, 16, 18, 22, 23, 26, 28, 30, 31, 32, 36, 37, 39, 40, 41, 43, 44, 46, 48, 50, 53, 54, 59, 60, 61, 62, 150, 151, 153, 154, 155, 156, 157, 158, 159, 160, 161, 162, 180, 181, 182, 183, 184, 187, 273, 275, 276, 277, 278, 279, 285, 293, 307, 310, 312, 313, 329, 333, 334, 335, 336, 337, 338, 339, 343, 346, 351, 354, 357, 375, 377, 444, 446, 448, 450, 457, 458, 459, 461, 462, 463, 464, 465, 467, 468, 469, 472, 475, 476, 477, 479, 480, 679, 682, 683.
\end{quote}

\begin{table}[t]
\centering
\caption{Dataset complexity for O2PL wins and losses against each baseline. Entries report the mean of $dK$ with the number of datasets in parentheses. Significance uses a seed-level sign test with a minimum mean reward gap of $0.02$.}
\label{tab:o2pl_dk_sig}
\begin{tabular}{lcc}
\toprule
Baseline & O2PL significant wins: mean $dK$ (count) & O2PL significant losses: mean $dK$ (count) \\
\midrule
Greedy & 421.16 (43) & 152.24 (33) \\
Cover-nu & 508.50 (30) & 217.56 (34) \\
Bag-greedy & 232.24 (25) & 245.65 (20) \\
RegCB-opt & 420.75 (57) & 26.00 (4) \\
\bottomrule
\end{tabular}
\end{table}

\subsection{The benefit of pessimism}
\label{subsec:offline_obp}
We also study O2PL for policy learning from logged bandit feedback, where the learner must construct a policy from a fixed batch of contextual bandit data. This setting highlights the benefit of pessimism under partial feedback.

\begin{figure}[t]
    \centering
    \begin{minipage}[t]{0.49\linewidth}
        \centering
        \includegraphics[width=\linewidth]{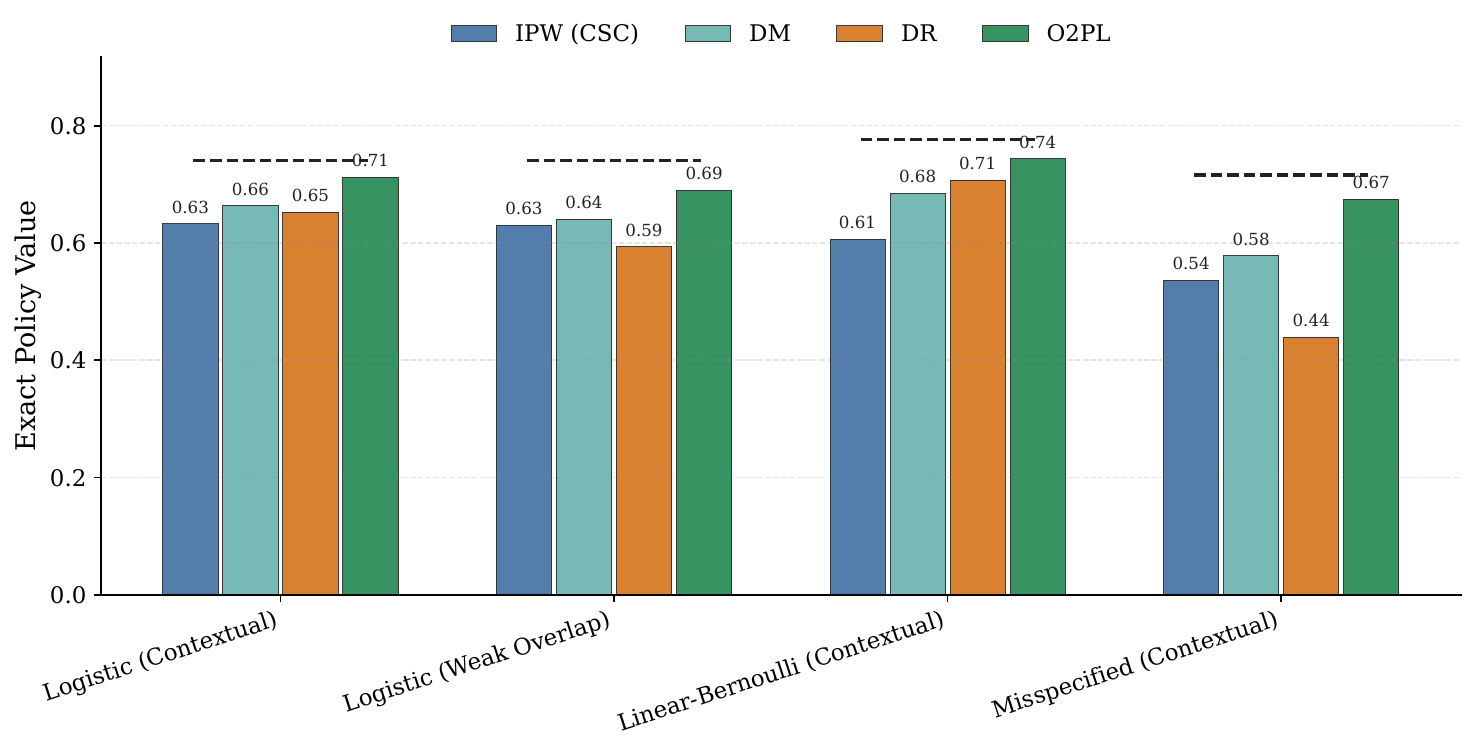}
        \caption{Exact policy value for policy learning on synthetic contextual bandit environments from logged data. The tasks vary the reward structure and the degree of overlap, including logistic, weak-overlap, linear Bernoulli, and misspecified settings. }
        \label{fig:offline_contextual_synthetic}
    \end{minipage}\hfill
    \begin{minipage}[t]{0.49\linewidth}
        \centering
        \includegraphics[width=\linewidth]{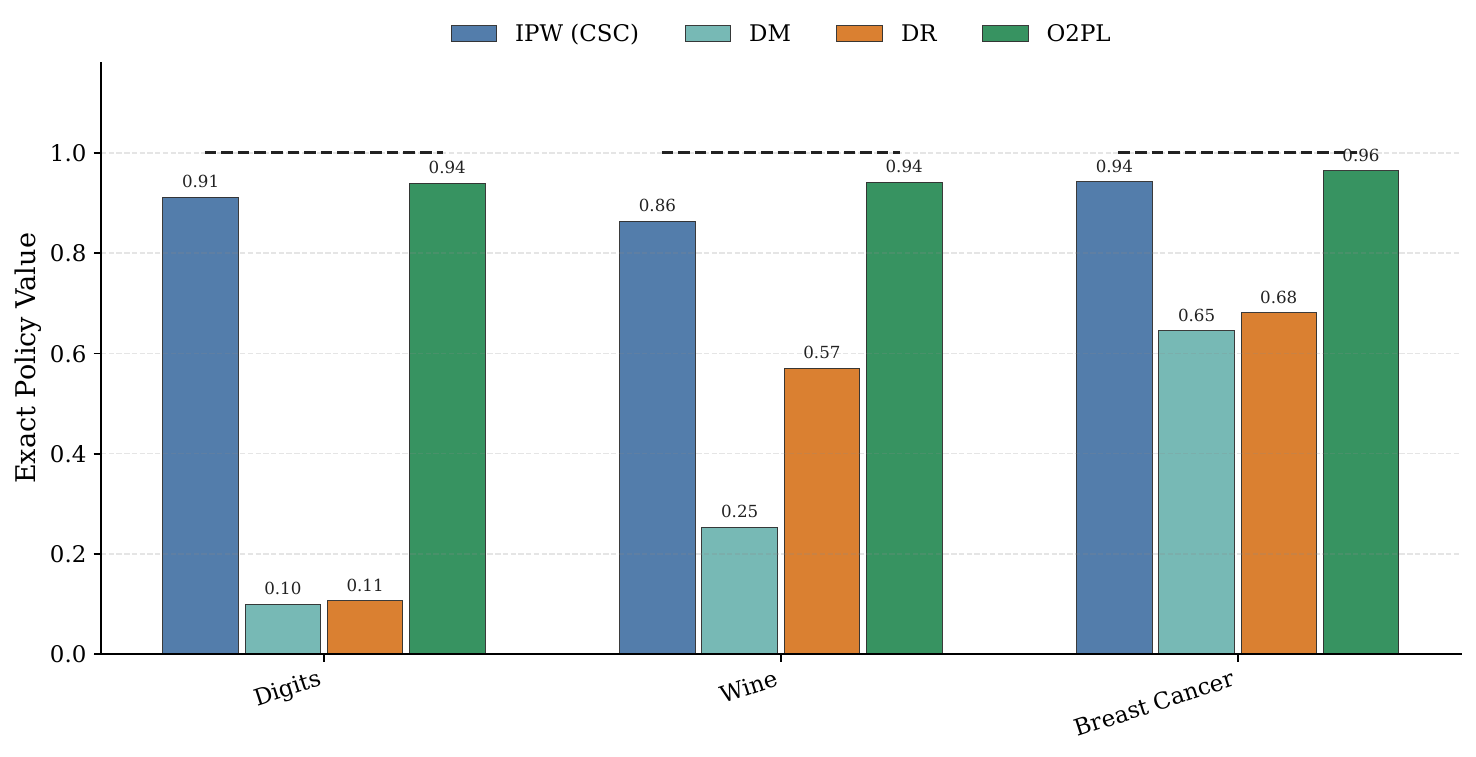}
        \caption{Exact policy value for policy learning on multiclass-to-bandit reductions from logged data. Each input is treated as a context, actions correspond to class labels, and rewards are binary. We compare the same baselines against O2PL. Dashed horizontal lines indicate the oracle policy value.}
        \label{fig:offline_multiclass_reduction}
    \end{minipage}
\end{figure}

We use the Open Bandit Pipeline (OBP) library \cite{saito2021openbanditdatasetpipeline}, which provides both synthetic contextual bandit tasks (Figure~\ref{fig:offline_contextual_synthetic}) and multiclass-to-bandit reductions (Figure~\ref{fig:offline_multiclass_reduction}).

We compare O2PL with standard policy learning baselines in OBP: IPW with a cost-sensitive classification reduction (\textsc{IPW-CSC}), a direct method (\textsc{DM}), and a doubly robust method (\textsc{DR}). These baselines span importance-weighted, model-based, and doubly robust approaches to learning from logged bandit data.

All methods are trained on the same logged sample, and we report exact policy value. As shown in Figure~\ref{fig:offline_contextual_synthetic}, O2PL is especially strong under weak overlap and misspecification, where pessimistic policy learning is most useful. Dashed horizontal lines indicate the oracle policy value.

\subsection{Additional numerical results}

\begin{figure*}[h]
  \centering
  \includegraphics[width=0.32\textwidth]{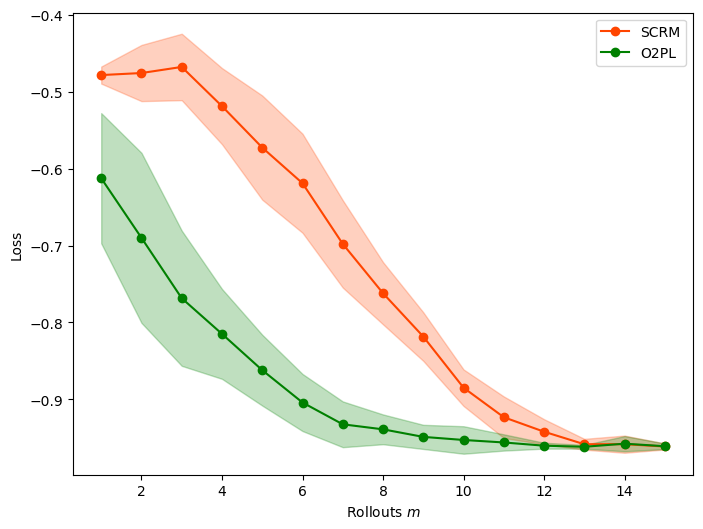}\hfill
  \includegraphics[width=0.32\textwidth]{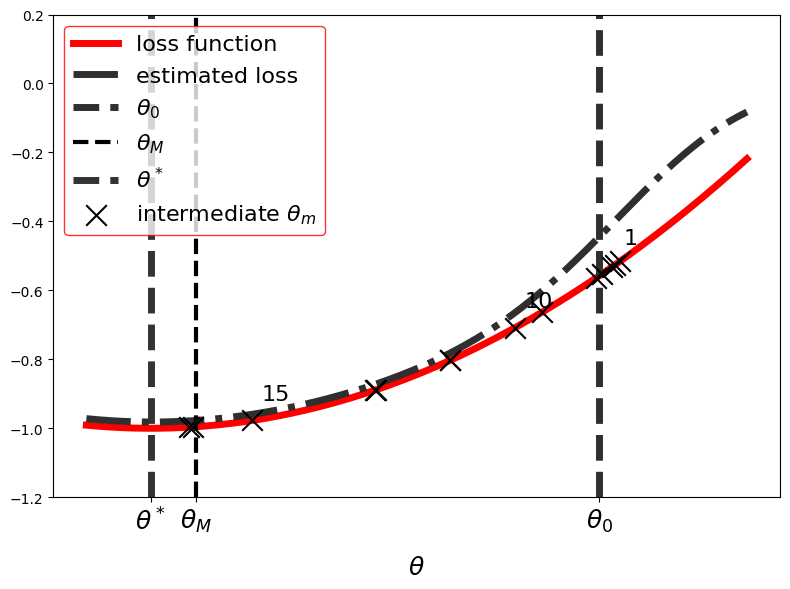}
  \includegraphics[width=0.32\textwidth]{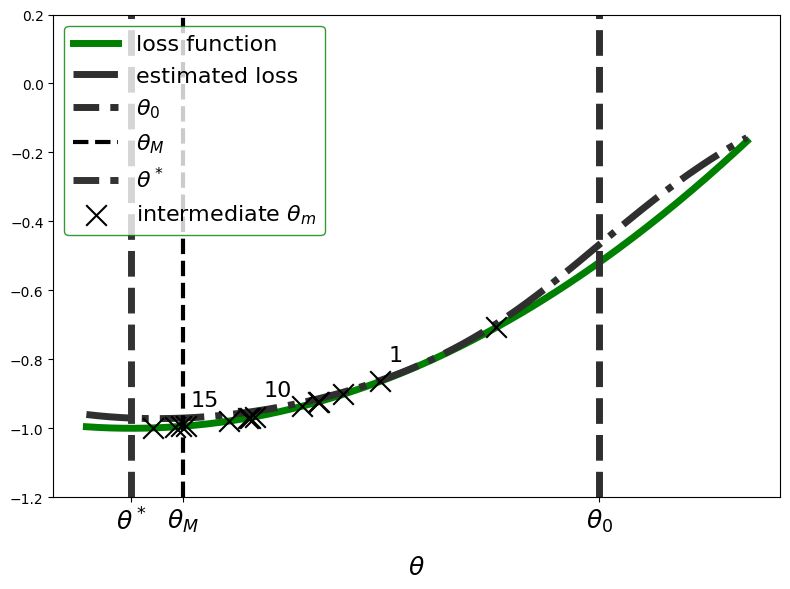}\hfill
  \caption{Loss evolution and optimization paths for (a) SCRM, (b) O2PL surrogate optimization, and (c) O2PL variance-penalized optimization.}
  \label{fig:all_four}
\end{figure*}

We now study the relative performance of our online learning algorithm O2PL compared to the geometric batch approach in SCRM \citep{zenati23scrm}. To illustrate the differences between our O2PL and SCRM, we set the following example.
Gaussian policies $\pi_\theta=\mathcal{N}(\theta,\sigma^2)$ incur loss
$l_t(a)=(a-y_t)^2-1$ with $y_t\sim\mathcal{N}(\theta^*,\sigma^2)$.
Figure~\ref{fig:all_four} reports loss over 15 rollouts for:
(i) \emph{SCRM}, trained adaptively with data from the last model $\pi_{m-1}$; and
(ii) \emph{O2PL}, trained on data from all deployed models $\pi_0,\dots,\pi_{m-1}$. O2PL stabilizes the training and improves learning.

To further compare our method to SCRM \citep{zenati23scrm}, we evaluate on both discrete and continuous action spaces and experiment in the batch setting.

\begin{table*}[h]
\centering
\caption{Loss values at $\lambda=\hat{\lambda}$ under \textbf{Uniform} and \textbf{Doubling trick}. Best (lowest) per environment within each block is in \textbf{bold}.}
\label{tab:loss_uniform_doubling_transposed}
\footnotesize
\setlength{\tabcolsep}{5pt}
\renewcommand{\arraystretch}{0.95}
\begin{adjustbox}{max width=\textwidth}
\begin{tabular}{l l c c c c c}
\toprule
\multicolumn{2}{c}{} &
\textbf{Pricing} &
\textbf{Advertising} &
\textbf{Toy\_env} &
\textbf{Syn-5} &
\textbf{Syn-20} \\
\midrule
\multirow{2}{*}{Uniform}
 & SCRM
 & $-5.355 \pm 0.196$
 & $-0.692 \pm 0.026$
 & $-0.454 \pm 0.002$
 & $0.534 \pm 0.098$
 & $0.451 \pm 0.059$ \\
 & O2PL
 & $\mathbf{-5.419 \pm 0.269}$
 & $\mathbf{-0.714 \pm 0.003}$
 & $\mathbf{-0.457 \pm 0.002}$
 & $\mathbf{0.230 \pm 0.039}$
 & $\mathbf{0.301 \pm 0.036}$ \\
\midrule
\multirow{2}{*}{Doubling}
 & SCRM
 & $-5.184 \pm 0.406$
 & $-0.676 \pm 0.048$
 & $-0.458 \pm 0.003$
 & $0.570 \pm 0.162$
 & $0.755 \pm 0.109$ \\
 & O2PL
 & $\mathbf{-5.568 \pm 0.036}$
 & $\mathbf{-0.717 \pm 0.002}$
 & $\mathbf{-0.458 \pm 0.003}$
 & $\mathbf{0.226 \pm 0.036}$
 & $\mathbf{0.479 \pm 0.193}$ \\
\bottomrule
\end{tabular}
\end{adjustbox}
\end{table*}

\paragraph{Toy.}
Let $d=10$. Contexts are iid $X\sim\mathcal N(0,I_d)$ and the action space is $\mathcal A=\mathbb R$.
The environment has a fixed weight $w_{\mathrm{env}}\in\mathbb R^d$; given $(X,a)$ the stochastic reward is
\[
R \;=\; (X^\top w_{\mathrm{env}})\,a \;-\; a^2 \;+\; \varepsilon,\qquad \varepsilon\sim\mathcal N(0,1)\ \text{independent}.
\]
Hence $\E[R\mid X,a]=(X^\top w_{\mathrm{env}})a-a^2$ is concave in $a$ with unique maximizer
\[
a^\star(X)=\tfrac12\,X^\top w_{\mathrm{env}},
\qquad
\E\big[R\mid X,a^\star(X)\big]=\tfrac14\,(X^\top w_{\mathrm{env}})^2.
\]
In the implementation, the environment returns \emph{targets} $t(X)=a^\star(X)$ and logs actions with a Gaussian behavior policy
\[
A\;\sim\;\mathcal N\!\big(t(X),\,\sigma_0^2\big)\quad\text{with}\ \ \sigma_0=\texttt{start\_std}=0.5,
\]
so propensities are the corresponding Normal pdf values at the realized $A$. The learning loss is the negated reward, capped at $0$ as in the code:
\[
\ell(A,X)\;=\;\min\!\Big\{-\big((X^\top w_{\mathrm{env}})A-A^2+\varepsilon\big),\ 0\Big\}.
\]

\paragraph{Pricing.}
As in \citep{demirer2019semi}, we use a personalized-pricing setup where a price $p$ must be chosen from context $x$ to maximize
\[
r(x,p)\;=\; p\big(a(x)-b(x)\,p\big)+\varepsilon,
\qquad \varepsilon\sim\mathcal N(0,1).
\]
Think of $d(x,p)=a(x)+b(x)p+\varepsilon$ as an unknown, context-specific demand term.
Contexts are $x\in[1,2]^k$ with $k>1$; only the first $\ell<k$ features affect demand (if $x=(z_1,\ldots,z_\ell,\ldots)$ we write $\bar x=(z_1,\ldots,z_\ell)$).
The logging policy is Gaussian, $p\sim\mathcal N(\bar x,1)$, centered at $\bar x$.
We use the quadratic specification $a(x)=2x^2$ and $b(x)=0.6x$ as in the original reference.

\paragraph{Advertising.}
As in \citep{zenati_counterfactual}, contexts $x\in\mathbb{R}^2$ are generated with \texttt{make\_moons} (noise $0.05$), yielding two user groups $g\in\{0,1\}$.
Each user has a latent responsiveness (target) $p_x\ge 0$ drawn from a group-specific Gaussian:
\[
p_x\mid g=0\sim \mathcal N(\mu_0,\,\sigma_p^2),\quad
p_x\mid g=1\sim \mathcal N(\mu_1,\,\sigma_p^2),\qquad
(\mu_0,\mu_1)=(3,1),\ \ \sigma_p=0.5,
\]
followed by an absolute value to enforce nonnegativity ($p_x:=|p_x|$).
Given a bid $a\in\mathbb{R}_+$, the one-step \emph{reward} is
\[
r(p_x,a)=\max\!\Bigg(\,
\begin{cases}
\displaystyle \frac{a}{p_x}, & a<p_x,\\[3pt]
1+\tfrac{1}{2}(p_x-a), & a\ge p_x,
\end{cases}\ ,\ -0.1\Bigg),
\]
and the learning loss used in experiments is the negated reward, $\ell(a,p_x)=-r(p_x,a)$.
The logging policy is Gaussian,
\[
a \sim \mathcal N(\mu_\ell,\sigma_\ell^2),\qquad \mu_\ell=2.0,\ \ \sigma_\ell=0.3,
\]
with propensities computed from the exact Normal density.

\paragraph{Discrete actions.}
We construct a contextual bandit from a multi-class dataset \(\{(x_i,l_i)\}_{i=1}^n\) by fitting a model \(g_\theta(x) \in \mathbb{R}^K\) and defining action probabilities via softmax:
\[
\pi_\theta(a\mid x) = \operatorname{softmax}(g_\theta(x))_a,\quad a\in \{1,\ldots,K\}.
\]
For each context \(x_i \in \mathbb{R}^d\), we sample \(A_i \sim \pi_\theta(\cdot\mid x_i)\), observe binary loss \(Y_i = \mathbf{1}\{A_i \neq l_i\}\), and record propensity \(p_i = \pi_\theta(A_i\mid x_i)\), yielding bandit data \(\mathcal{D}_{\text{bandit}} = \{(X_i, A_i, Y_i, p_i)\}_{i=1}^n\). We generate synthetic \textsc{Syn} data with \texttt{make\_multiclass} from \texttt{scikit-learn}. The number of actions is $3$ and we evaluate for context dimensions $d=5$ ('Syn-5') and $d=20$ ('Syn-20'). The logging policy $\pi_0$ is multinomial logistic regression (lbfgs, $L_2$, $C{=}1$, \texttt{max\_iter}=10000) trained on the first $10\%$ of $X_{\text{train}}$. For a batch $X$, it outputs class probabilities $P\in[0,1]^{n\times K}$, samples $A_i\sim\text{Cat}(P_{i,\cdot})$, and returns propensities $w_i=P_{i,A_i}$. $\pi_\theta$ is linear-softmax: logits $=XW$ with $W\in\mathbb{R}^{d\times K}$, probabilities via softmax, actions sampled by inverse-CDF; propensities $\pi_\theta(A\mid X)$ returned exactly. Greedy online evaluation uses $\arg\max_a$ logits and 0--1 loss on the test split.

Table~\ref{tab:loss_uniform_doubling_transposed} reports results for both continuous and discrete action settings. To strengthen our comparison with SCRM, we consider two sampling regimes: a fixed batch size (\textit{uniform}) and exponentially increasing batch sizes (\textit{doubling trick}). Across all environments, O2PL consistently matches or outperforms SCRM, highlighting the benefits of fully online updates.

Experiments were run on an average of 20 different seeds. For the discrete-action experiments, with the doubling trick we used $n_0=240$, $M=5$, and an implementation floor parameter $\tau=0.001$; with uniform splits we used $n_0=1500$, $M=5$, and the same $\tau=0.001$.

For the continuous-action environments, experiments were run on an average of 20 different seeds. With the doubling trick we used $n_0=150$, $M=10$, and an implementation floor parameter $\tau=0.001$; with uniform splits we used $n_0=1500$, $M=10$, and the same $\tau=0.001$.

\end{document}